\documentclass[dvipsnames]{article} %
\usepackage{colm2024_conference}

\usepackage{booktabs}
\usepackage{enumitem}
\usepackage{wrapfig}
\usepackage{algorithm}
\usepackage{algpseudocode}
\usepackage{graphicx}
\usepackage[misc]{ifsym}
\usepackage{multicol} 
\usepackage{microtype}
\usepackage{colortbl}
\usepackage[utf8]{inputenc}
\usepackage[T1]{fontenc}
\definecolor{lightgray}{rgb}{0.9,0.9,0.9}
\usepackage{caption}
\usepackage{subcaption}
\usepackage{graphicx}
\usepackage{dsfont}
\usepackage{setspace}
\usepackage{url}
\usepackage{multirow}
\usepackage{tabularx}
\usepackage{blindtext}
\usepackage{pgfplots}
\pgfplotsset{compat=1.18} 
\usepackage{tikz}
\usetikzlibrary{er,positioning,bayesnet}
\usepackage{makecell}
\usepackage{tipa}
\usepackage{siunitx}
\usepackage{nicefrac}
\usepackage{listings}
\usepackage[raster,skins, most]{tcolorbox} %
\usepackage{xltabular}
\usepackage{adjustbox}
\usepackage{xurl}
\usepackage{rotating}
\usepackage[normalem]{ulem}
\usepackage{graphicx}

\usepackage{amsthm}

\theoremstyle{plain}
\theoremstyle{definition}

\theoremstyle{remark}

\usepackage{placeins}

\usepackage{fontawesome}

\usepackage[table]{xcolor}

\usetikzlibrary{calc}

\useunder{\uline}{\ul}{}

\usepackage{amsmath,amsfonts,bm}

\def\eqref#1{equation~\ref{#1}}
\def\1{\bm{1}}

\DeclareMathAlphabet{\mathsfit}{\encodingdefault}{\sfdefault}{m}{sl}
\SetMathAlphabet{\mathsfit}{bold}{\encodingdefault}{\sfdefault}{bx}{n}

\newcommand*\justify{%
  \fontdimen2\font=0.4em% interword space
  \fontdimen3\font=0.2em% interword stretch
  \fontdimen4\font=0.1em% interword shrink
  \fontdimen7\font=0.1em% extra space
  \hyphenchar\font=`\-% allowing hyphenation
}

\renewcommand{\texttt}[1]{%
  \begingroup
  \ttfamily
  \begingroup\lccode`~=`/\lowercase{\endgroup\def~}{/\discretionary{}{}{}}%
  \begingroup\lccode`~=`[\lowercase{\endgroup\def~}{[\discretionary{}{}{}}%
  \begingroup\lccode`~=`.\lowercase{\endgroup\def~}{.\discretionary{}{}{}}%
  \catcode`/=\active\catcode`[=\active\catcode`.=\active
  \justify\scantokens{#1\noexpand}%
  \endgroup
}

\usepackage{makecell}
\usetikzlibrary{tikzmark}
\makeatletter
\newcommand*\myfontsize{%
  \@setfontsize\myfontsize{7}{8}%
}
\makeatother

\definecolor{uclablue}{RGB}{159, 195, 224}

\definecolor{uclagold}{RGB}{255, 240, 180}

\definecolor{aliceblue}{RGB}{255, 238, 241}

\definecolor{cadmiumgreen}{rgb}{0.0, 0.42, 0.24}

\definecolor{myred}{rgb}{0.7, 0.3, 0.0}
\definecolor{myblue}{rgb}{0.2, 0.3, 0.6}
\definecolor{babygreen}{rgb}{0.85, 0.97, 0.85}

\definecolor{purple1}{RGB}{126, 107, 196}
\definecolor{purple2}{RGB}{199, 158, 207}
\definecolor{purple3}{RGB}{214, 200, 255}
\definecolor{purple4}{RGB}{254, 240, 255}

\definecolor{deepblue}{RGB}{48, 58, 82}

\definecolor{SoftLavender}{HTML}{F3EEFF}

\definecolor{deepPurple}{HTML}{330066}
\definecolor{uclablue_old}{rgb}{0.15, 0.45, 0.68}
\hypersetup{
    breaklinks,
    citecolor=uclablue_old,
    colorlinks=true,
}
\usepackage[para]{footmisc}

\definecolor{basegray}{RGB}{160,165,175}
\definecolor{basegraydark}{RGB}{110,115,125}
\definecolor{rlblue}{RGB}{47,128,237}
\definecolor{rlbluefill}{RGB}{220,235,255}

\usetikzlibrary{arrows.meta,calc,positioning,backgrounds}

\newcounter{daggerfootnote}

\title{%
\begin{tabular}[t]{l} 
  \parbox[t]{0.8\textwidth}{\centering 
    Beyond the Best Guess: Improving LLM Solution Coverage with Evolution Strategies
  }
\end{tabular}
}

\author{%
\large 
Conor F. Hayes$^{1}$, 
Elliot Meyerson$^{1}$, 
Kajetan Schweighofer$^{1}$, 
Roberto Dailey$^{1}$, 
Babak Hodjat$^{1}$, 
Risto Miikkulainen$^{1,2}$, 
Xin Qiu$^{1}$
  \\[1em]
\normalsize
$^{1}$Cognizant AI Lab, San Francisco \quad $^{2}$The University of Texas at Austin, Austin
}

\newcommand{\hflogo}{%
  \raisebox{-0.2ex}{\includegraphics[height=1.1em]{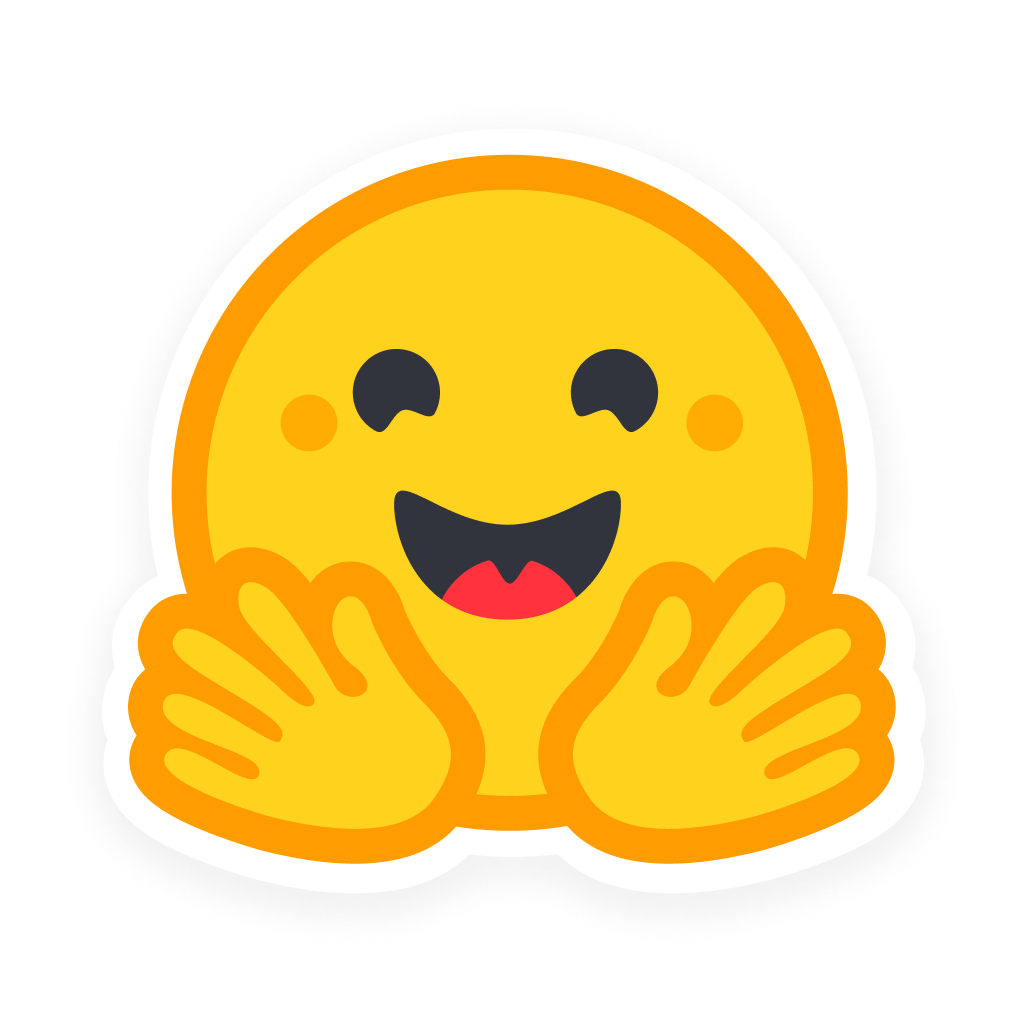}}%
}
\newcommand{\hflink}[1]{%
  \hflogo\hspace{3pt}\href{https://huggingface.com/collections/conorfhayes/#1}{\texttt{#1}}%
}

\begin{document}

\maketitle

\begin{abstract}
Large Language Models (LLMs) are increasingly deployed in discovery domains such as math and science. The usual approach is to present the problem to the model and use its answer as the proposed solution. However, beyond this best guess, discovery can be enhanced by increasing test-time compute. In a process called pass@k, the model is allowed to explore the solution space and generate diverse candidate solutions. Unfortunately, the standard approach to post-training LLMs through Reinforcement Learning (RL) may limit pass@k: the model's output distribution narrows around high-reward outputs, causing the solution coverage to collapse. The alternative is to use Evolution Strategies (ES), a population-based, gradient-free post-training method that optimizes directly in weight space through random perturbations. As this paper shows, ES achieves consistently higher pass@k than RL and produces a broader output distribution with greater solution coverage. This coverage in turn makes it possible to achieve better results in e.g.\ standard math benchmarks. Thus, ES provides a better foundation for post-training in discovery problems and other domains where diverse solution coverage is critical. 

\begin{center}
\faGithub{} \hspace{2pt}
\href{https://github.com/conorfhayes/beyond-the-best-guess}{\texttt{beyond-the-best-guess}}
\\
\hflink{beyond-the-best-guess}

\end{center}
\end{abstract}

\begin{figure}[h]                  % use figure* for full width in a 2-column paper
  \centering
  \begin{subfigure}{0.4\linewidth}
    \centering
    \includegraphics[width=\linewidth]{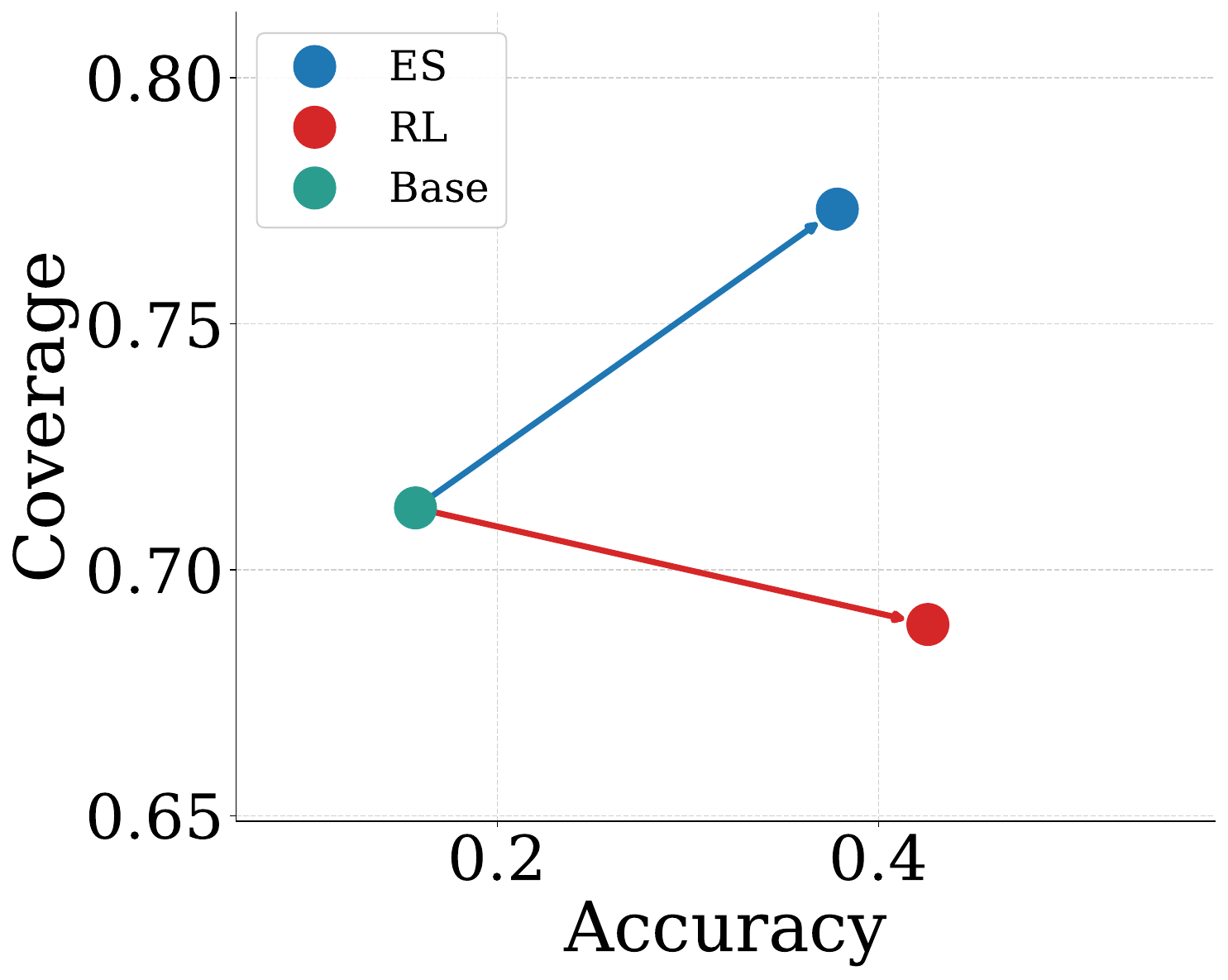}
    %\caption{}
    \label{fig:teaser-arrow}
  \end{subfigure}
  \hspace{0.04\textwidth}
  \begin{subfigure}{0.4\linewidth}
    \centering
    \includegraphics[width=\linewidth]{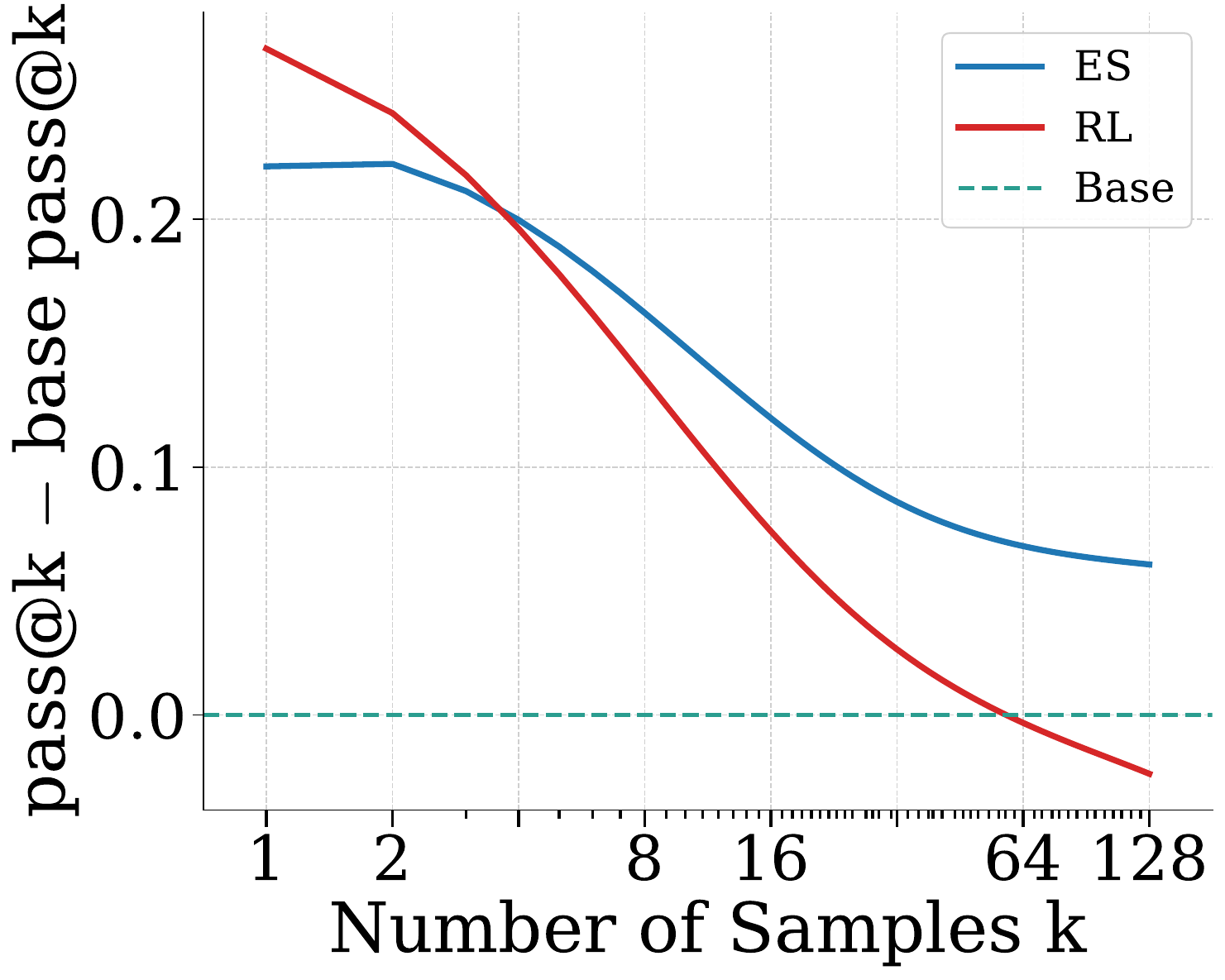}
    %\caption{}
    \label{fig:teaser-delta}
  \end{subfigure}
  \caption{\textbf{Whereas RL sharpens accuracy, ES broadens solution coverage}. (Left) Accuracy--coverage shift. With respect to the base model, RL raises accuracy (pass@1) but lowers coverage (pass@k), whereas ES raises both. (Right) Coverage relative to the base model vs.\ the number of samples ($k$). ES stays positive across all measured $k$ values (max $k{=}128$). In contrast, as $k$ increases RL reduces coverage below ES, and eventually falls even below the base model.}
  \label{fig:teaser}
\end{figure}

\newpage

\section{Introduction}
\label{sec:intro}

The reasoning capabilities of Large Language Models (LLMs) continue to improve \citep{team2025kimi,zeng2026glm} and, as a result, LLMs are increasingly being utilized for discovery problems, e.g., where the model is tasked with solving open scientific problems in math or coding. The continued improvement in reasoning capabilities is largely due to the introduction of post-training, where Reinforcement Learning \citep[RL;][]{sutton1998reinforcement}, is utilized in settings where model responses can be verified with outcome based verifiable rewards \citep[RLVR;][]{lambert2024tulu}. RLVR training increases model performance across a broad range of tasks, including math and coding. This has become a crucial step in the LLM training pipeline.

The increased reasoning capabilities of LLMs, combined with the ability to scale test-time compute, have enabled these models to tackle open scientific problems that were previously the sole domain of human experts \citep{lu2026towards}. For example, OpenAI recently disproved an 80-year-old geometry conjecture by finding a point arrangement that no human mathematician had discovered \citep{openai2026erdos}. This result required scaling test-time compute across many candidate proofs \citep{openai2026erdosblog}. By scaling test-time compute, a model is able to sample many feasible solutions to a given problem, effectively performing search over the solution space \citep{snell2024scaling}. This shift in paradigm means traditional single-shot metrics no longer capture what matters, e.g. pass@1. Instead, a model's value increasingly lies not in whether its single best guess is correct, but in whether some sample among many contains a correct solution. Capturing this property requires a different measure of performance, pass@k \citep{brown2024large}, which quantifies solution coverage as the probability that at least one correct solution exists within $k$ samples drawn from the model's output distribution. Maximizing pass@k therefore requires the model's output distribution to maintain broad support over valid solutions.

Before test-time scaling, models have usually been refined through post-training with RL. It turns out that such RL-trained LLMs suffer from distribution collapse: RL updates sharpen the output distribution to maximize the best guess, i.e.\ pass@1, concentrating probability mass on high-reward outputs while narrowing the support, pruning low-probability but correct solutions \citep{nguyen2026reasoning}. This process reduces the overall solution coverage of a model and limits the models ability to explore alternative viable solutions for a given problem. Remarkably, even the base model can outperform an RL fine-tuned model at pass@k, a phenomenon first reported by \citet{yue2025does}. This mismatch suggests that RL may be poorly suited to post-training in domains where pass@k is the metric of interest.

Recently, Evolution Strategies (ES) has emerged as an alternative LLM post-training technique with potentially better fit for pass@k \citep{qiu2025evolution}. ES is a population-based gradient-free optimization method that optimizes model parameters through random perturbations in weight space \citep{salimans2017evolution}. Random perturbations are used to generate a population of models that are evaluated and rewarded on the basis of their performance in a given task. ES achieves comparable performance to RL in reasoning tasks; however, ES optimizes for a robust solution distribution \citep{lehman2018more} resulting in a more broad distribution of solutions for pass@k to utilize. 

This difference arises because ES and RL perform parameter updates differently. RL typically uses policy gradient methods \citep{williams1992simple,sutton1998reinforcement} such as Proximal Policy Optimization \citep[PPO;][]{schulman2017proximal} or Group Relative Policy Optimization \citep[GRPO;][]{shao2024deepseekmath} to update a single model's parameters using gradients computed over sampled token sequences. It operates in the action space, i.e.\ the gradient signal is used to increase the log-probability of high-reward outputs directly, progressively concentrating probability mass on a narrow set of solutions and sharpening the output distribution towards a single high-reward mode of the output distribution \citep{wu2025invisible,nguyen2026reasoning}.
This narrowing is further compounded in settings where the reward is sparse and binary \citep{sinha2026expected}, reinforcing a small subset of high-reward trajectories at the expense of the broader solution space \citep{yue2025does}. In contrast, ES operates in the parameter space: it maximizes the expected reward over a distribution of weight-space perturbations
%, exploring directly in weight space rather than through token-level gradient signals
\citep{salimans2017evolution,conti2018improving}.
As a result, the ES optimization pressure is not focused on sharpening a single high reward mode of the output distribution, but instead pushes the model parameters to regions robust to weight perturbations and comprising a broader set of solutions \citep{lehman2018more}. Recent work has also shown that ES fine-tuning results in smaller deviation from the base model's output distribution than RL on the training task \citep{schweighofer2026overcoming}, suggesting that ES post-training preserves broader output distribution support and makes it better suited for settings where solution diversity is critical. 

This paper empirically investigates whether ES yields improved pass@k relative to RL-trained models at matched compute budgets and evaluates the implications for downstream performance through test-time scaling experiments. The main contributions are:

\begin{itemize}
    \item The most comprehensive evaluation of ES as an LLM post-training technique to date: ES models are trained on different parameter scales from 1.5B to 32B parameters, spanning multiple model families.
    \item Establishing ES as a superior post-training strategy for settings requiring solution diversity: ES consistently improves pass@k over RL-trained models across all model families and scales.
    \item Insights into the mechanistic origins of the observed improvements in solution coverage, obtained through a systematic analysis of the output distributions of ES-trained models relative to base and RL-trained models. 
    \item Demonstrating the value of high-quality ES output distributions compared to those of RL in several math benchmarks.
\end{itemize}
Overall, the results confirm the hypothesis that ES is better suited to pass@k than RL, and establish ES as the state-of-the-art approach to post-training in domains where diverse solution exploration is critical to performance.

\section{Background}
This section reviews the literature most relevant to this work, covering four areas: Reinforcement Learning with verifiable rewards as the dominant LLM post-training paradigm, test-time scaling and its relationship to solution coverage, the limitations of Reinforcement Learning, and Evolution Strategies as an emerging alternative to Reinforcement Learning for LLM post-training.

\subsection{Reinforcement Learning with Verifiable Rewards}
Reinforcement Learning (RL) for LLMs uses policy gradient methods \citep{williams1992simple,sutton1998reinforcement} where gradient updates are computed using the log-probability of sampled token sequences and their corresponding reward.
\citet{shao2024deepseekmath} combined RL with verifiable rewards \citep[RLVR;][]{lambert2024tulu} to elicit chain-of-thought \citep[CoT;][]{wei2022chain} reasoning behaviors from LLMs.
In the RLVR setting, a deterministic verifier generates a reward of $1$ for a correct answer and $0$ for an incorrect answer.
A format reward may also be added to encourage the model to explicitly separate the reasoning process from the final answer.
The goal of RL is to find a policy $\pi_{\theta}$ that maximizes the expected reward :

\begin{equation}
    J(\theta) = \mathbb{E}_{x \sim \mathcal{D},\, y \sim \pi_{\theta}(\cdot \mid x)} \left[ R(y) \right],
\end{equation}

where $x$ is a problem sampled from a dataset $\mathcal{D}$, $y$ is a response sampled from the policy $\pi_{\theta}$, and $R(y)$ is the reward assigned by the verifier.
In practice, policy gradient methods such as GRPO \citep{shao2024deepseekmath} are used in RLVR settings \citep{zeng2025glm} to optimize $J(\theta)$ by computing gradient updates that increase the log-probability of high-reward responses. %As a result, this has led to many algorithmic enhancements of GRPO over time \citep{liu2025understanding,yu2026dapo,zheng2025group}, and
RLVR has become the standard post-training paradigm, demonstrating strong performance across mathematical reasoning, science discovery, and coding benchmarks \citep{olmo2025olmo,guo2025deepseek}.
%Once trained, the final LLM is deployed for many use cases. Recently, test-time scaling (TTS) \citep{snell2024scaling,zhang2025survey} has become a method by which inference time compute can be used to boost performance across many settings \citep{jaech2024openai}, e.g, scientific discovery \citep{openai2026erdos,feng2026aletheia}.

\subsection{Test-Time Scaling}
\label{subsec:related_work_tts}

The classical way to assess the quality of an LLM in reasoning domains is to look at single sample accuracy, i.e. pass@1. A single answer is drawn from the model and checked for correctness.
Over the past few years, researchers have realized that one way to increase model performance is to systematically increase the amount of compute spent at test time.
This regime of model usage is known as Test-time Scaling \citep[TTS;][]{zhang2025survey}.
TTS began with early breakthroughs like CoT \citep{wei2022chain}, and has since diversified into methods including tree-base approaches \citep{yao2023tree,aygun2026ai}, iterative feedback loops \citep{madaan2023self, shinn2023reflexion}, and general agentic harnesses \citep{ning2026code}.

TTS works because there is some latent knowledge in the LLM that may not be actualized when asking directly for a single answer, but increasing the amount of interaction with the model can bring the answer forward.
The most natural way to assess whether a model contains the desired knowledge or capabilities is pass@k, i.e.\ sampling $k > 1$ answers and checking whether any is correct.
In addition to its utility as an analytical tool, pass@k is a practical and easily parallelizable TTS method to improve performance in domains where answers are verifiable \citep{brown2024large, snell2024scaling}.
It can be further used in non-verifiable test settings by incorporating a mechanism of voting across answers \citep{wang2022self, meyerson2025solving}.
Section \ref{sec:pass@k-performance} focuses on pass@k in verifiable settings in order to characterize differences between the behavior of ES- and RL-tuned models. The voting results in Section~\ref{sec:test_time_scaling} verify that improvement in pass@k leads to downstream benefits in non-verifiable domains as well.

\subsection{Limitations of RLVR}

TTS is typically applied to deployed models that have undergone post-training with RLVR. Although RLVR post-training has shown improved pass@1 performance, further analysis of the output distributions of RL-trained models has revealed a number of fundamental limitations.

A key issue is distributional collapse, where the policy gradient updates in RL progressively concentrate probability mass on high-reward outputs, narrowing the support of the output distribution \citep{wu2025invisible}. This narrowing effect is particularly problematic in sparse, binary reward settings, where a small subset of high-reward trajectories are reinforced at the expense of the broader solution space. As a consequence, low-probability but correct solutions are effectively pruned from the distribution, reducing the overall solution coverage of the model \citep{nguyen2026reasoning,wu2025invisible}. This reduction in coverage makes TTS less effective because the model is unable to explore diverse candidate solutions regardless of the compute budget allocated \citep{dang2025weight}.

A striking manifestation of this effect is that for sufficiently large $k$, the pass@k accuracy of the base model can surpass that of the RLVR fine-tuned model \citep{yue2025does}. This counterintuitive result demonstrates that RL post-training can be actively harmful in TTS settings, trading solution coverage for pass@1 performance in a way that becomes increasingly costly as $k$ grows.

Several methods have been proposed to mitigate distribution collapse in RL. KL-divergence penalties \citep{schulman2017proximal} constrain the fine-tuned model to remain close to the reference distribution, preventing the policy from straying too far from its initialization. Another distributional collapse mitigation strategy is the use of resets \citep{liu2026prorl,bartoldson2026trajectory} where the reference distribution is periodically updated to the current policy during training, relaxing the KL constraint over time to avoid over-constraining exploration \citep{wu2025invisible}. A complementary approach addresses the clipping threshold in the policy gradient objective. A tight upper clip limits the policy update step, restricting exploration and contributing to distribution collapse. \citet{yu2026dapo} address clipping restriction by increasing the upper clipping threshold, allowing the policy to make larger updates and to explore a broader region of the solution space. Other approaches address distribution collapse by modifying the policy gradient objective directly. Vector Policy Optimization \citep[VPO;][]{bahlous2026vector} reformulates RL as a multi-objective RL problem \citep{hayes2022practical}, training the model to produce candidate sets that cover the Pareto frontier of a vector-valued reward, thereby preserving diversity in the output distribution. While VPO improves pass@k, it does so at the cost of pass@1 performance.

However, these approaches introduce additional hyperparameters that are difficult to tune and may conflict with the primary reward objective, potentially making them less effective in practice \citep{shah2025comedy}. Furthermore, these approaches do not fundamentally address the tension between policy gradient optimization and distributional diversity. This motivates the investigation of alternative post-training methods that preserve solution coverage by construction rather than through explicit regularization.

\subsection{Evolution Strategies for LLM Fine-tuning}
Evolution strategies (ES) are a class of population-based zeroth-order optimization methods \citep{wierstra2014natural,rechenberges1973}. Instead of backpropagation, ES approximates a gradient using random weight perturbations over a population $N$. For a given set of parameters $\theta$ and a reward function $R(\cdot)$, at each optimization step, for each individual in the population $n \in N$, ES samples perturbations $\epsilon_{n} \sim \mathcal{N}(\bm{0}, I)$, scales them by $\sigma$, and evaluates the performance of the perturbed model $r_{n} = R(\theta + \sigma \epsilon_n)$ under the reward function. Using this approach, ES  aims to maximize the expected reward over the population under Gaussian perturbations: $\max_{\theta} \mathbb{E}_{\epsilon_{n} \sim \mathcal{N}(\bm{0}, I)} \left[ R(\theta + \sigma \epsilon) \right]$, where the reward is normalized over the population using ranking \citep{salimans2017evolution} or z-scores \citep{qiu2025evolution}. Finally, the weight update aggregates the perturbations weighted by their normalized reward as
\begin{equation} \label{eq:es}
    \theta_{t} = \theta_{t - 1} + \alpha \cdot \frac{1}{N} \sum_{n=1}^{N} \hat{r}_{n} \epsilon_{n} \ ,
\end{equation}
where $\alpha$ is the learning rate and $\hat{r}_{n}$ is the normalized rewards for each member of the population.

\citet{salimans2017evolution} implemented ES to optimize a policy in RL domains showing competitive performance with RL. Given that ES does not compute a gradient and therefore no backpropagation is required, they highlighted how ES can be more scalable and parallelizable than RL. As the number of model parameters grows, computing gradients becomes more computationally difficult. 

Recently, \citet{qiu2025evolution} applied ES to LLM fine-tuning, showing competitive performance with RL across many domains such as mathematical reasoning. 
Interestingly, \citet{qiu2025evolution} used a population size of $30$, breaking the long-held assumption that population sizes needed to be on the order of thousands for ES to optimize effectively. 
As a result, ES has become an attractive alternative to RL, motivating further algorithmic studies \citep{liang2026blessing,hoy2026matching} and new post-training methods \citep{sarkar2025evolution,gan2026neural,schweighofer2026overcoming,xu2026quantized}. This work contributes to this growing body of research by empirically investigating the solution coverage of ES-trained models, comparing pass@k performance against state-of-the-art RL-trained LLM checkpoints across mathematical reasoning benchmarks.

\section{Experimental Setup}

This section describes the experimental setup for comparing ES and RL fine-tuning on pass@k across multiple benchmarks and models ranging from 1.5B to 32B parameters.

%ES and RL fine-tuning methods were compared on pass@k across a broad set of mathematical reasoning benchmarks and models ranging from 1.5B to 32B parameters.

\subsection{The Pass@k Metric}
Given an input prompt 
%and a fixed compute budget $k$, 
the model samples $k$ responses, and the task is considered solved if at least one of the $k$ responses is correct. Naively measuring pass@k requires many trials for each $k$ to reduce estimation variance.
To address this, \citet{chen2021evaluating} introduced an unbiased low variance pass@k estimator over a dataset $\mathcal{D}$:
\begin{equation}
\label{eqn:pass@k}
\text{pass}@k = \mathbb{E}_{x \sim \mathcal{D}} \left[ 1 - \frac{\binom{n-c}{k}}{\binom{n}{k}} \right],
\end{equation}
where $n$ is the total number of sampled responses per problem and $c$ is the number of correct responses for problem $x \in \mathcal{D}$. As $k$ increases, pass@k monotonically increases toward the fraction of problems for which at least one correct solution exists in the model's output distribution, providing a measure of solution coverage. At $k=1$, pass@k reduces to the standard accuracy metric, while for large $k$ it captures the breadth of the model's output distribution rather than the quality of its single most likely response. 
% As a result, both accuracy and coverage are accurately represented with Equation. \ref{eqn:pass@k}. 
For each benchmark below, the same $n$ is used as in \citet{yue2025does}.

%Similarly to \citet{yue2025does}, for each benchmark, $n$ is chosen to maximize performance and changes per benchmark. Thus for GSM8K, Olympiad Bench, and Minerva $n=128$, while for MATH500 $n=256$. 

\subsection{The Mathematical Reasoning Domain}
\label{sec:pass@k-math}

To study pass@k behavior, standard math reasoning benchmarks and training setups were used.

\paragraph{Benchmarks.} To study pass@k for mathematical reasoning two datasets are used for training: GSM8K \citep{cobbe2021training} and MATH (level 3-5) \citep{hendrycks2021measuring,liu2025understanding}. Each dataset captures a different level of problem difficulty, allowing the effects of ES and RL on solution coverage to be tested across easier and harder reasoning tasks. Models trained using GSM8K are evaluated using the GSM8K test set \citep{cobbe2021training}, while models trained using MATH are evaluated using MATH500 \citep{hendrycks2021measuring}, Olympiad Bench \citep{he2024olympiadbench}, and Minerva \citep{lewkowycz2022solving}. As in prior work, at test time each response is limited to 16,384 tokens which are sampled with temperature $0.6$ and top-p $0.95$ \citep{yue2025does}. %All further ES and RL evaluation details are documented in Appendix Y.

\paragraph{Training.} Pass@k performance for ES is extensively evaluated on GSM8K by fine-tuning models across sizes and families: Qwen2.5 1.5B-Instruct, 3B-Instruct, and 7B-Instruct \citep{qwen2025qwen25technicalreport}, and Qwen3 1.7B, 4B, and 8B \citep{yang2025qwen3}. The ES-at-Scale library \citep{qiu2025evolution} is used for ES and the VERL library \citep{sheng2025hybridflow} for RL. To study pass@k at larger scales, ES-at-Scale is used to fine-tune Qwen2.5-Math-7B, Qwen2.5-14B, and Qwen2.5-32B \citep{qwen2025qwen25technicalreport} on MATH \citep{hendrycks2021measuring}, comparing to state-of-the-art publicly available RL checkpoints from SimpleRL-Zoo \citep{zeng2025simplerl} and OatZero \citep{liu2025understanding}.

\section{Pass@k Performance}
\label{sec:pass@k-performance}

This section presents results on comparing the pass@k performance of ES and RL across benchmarks and models.

\subsection{GSM8K Results}

\begin{figure}[t!]
    \centering

    \begin{subfigure}[b]{0.3\columnwidth}
        \centering
        \includegraphics[width=\textwidth]{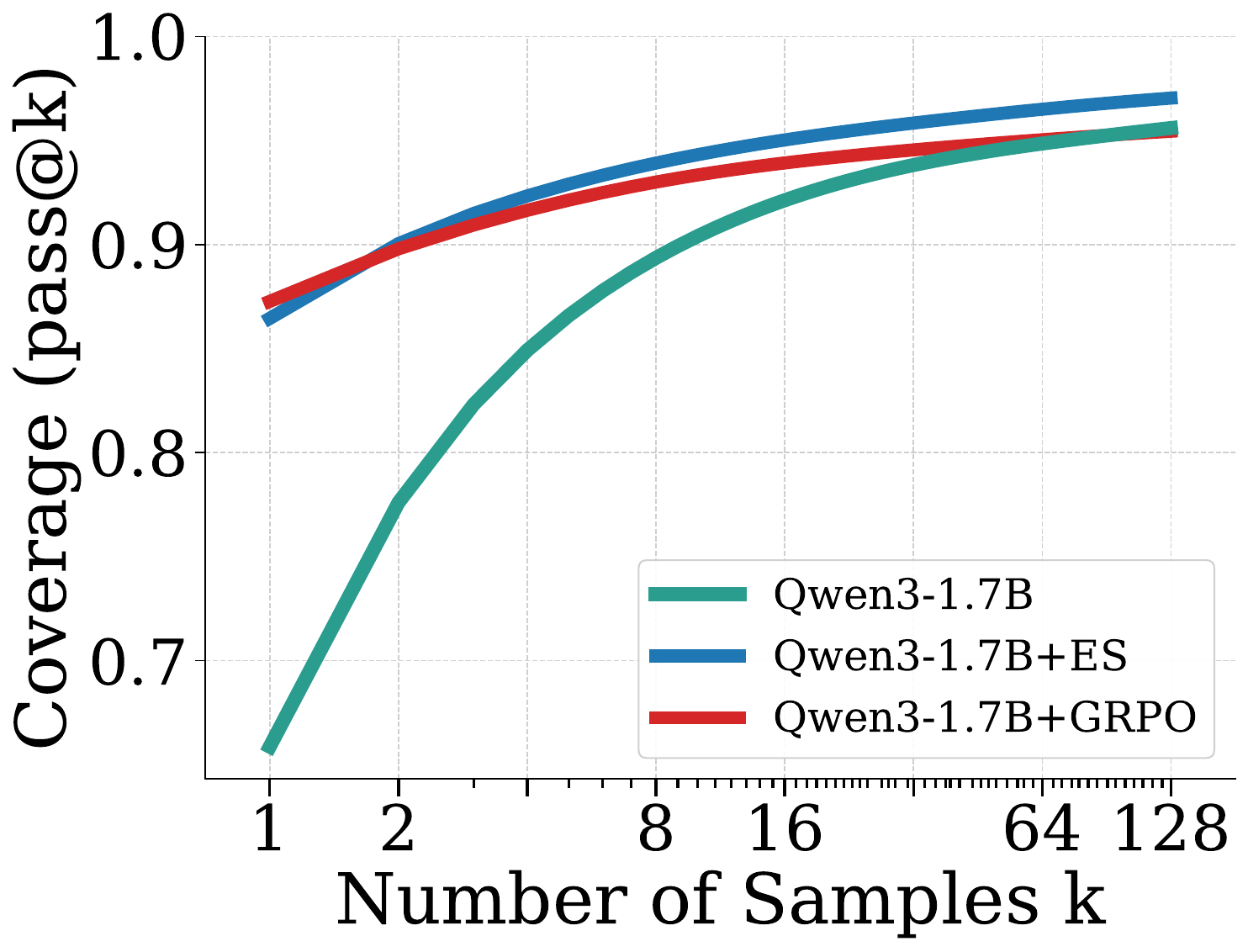}
        \caption{Qwen3-1.7B}
        \label{fig:qwen2.5-1.5b-instruct-gsm8k}
    \end{subfigure}
    \hfill
    \begin{subfigure}[b]{0.3\columnwidth}
        \centering
        \includegraphics[width=\textwidth]{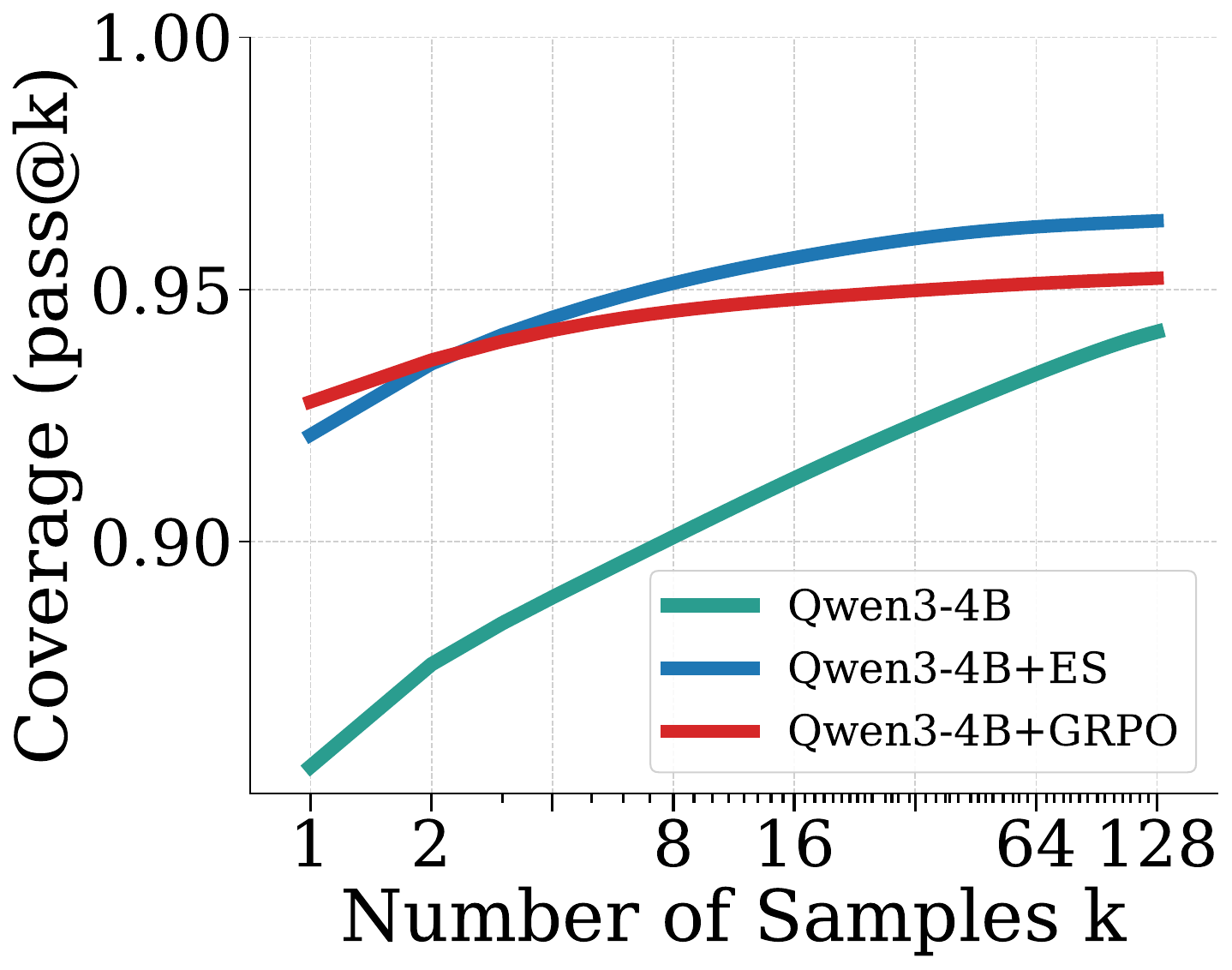}
        \caption{Qwen3-4B}
        \label{fig:qwen3-4b-gsm8k}
    \end{subfigure}
    \hfill
    \begin{subfigure}[b]{0.3\columnwidth}
        \centering
        \includegraphics[width=\textwidth]{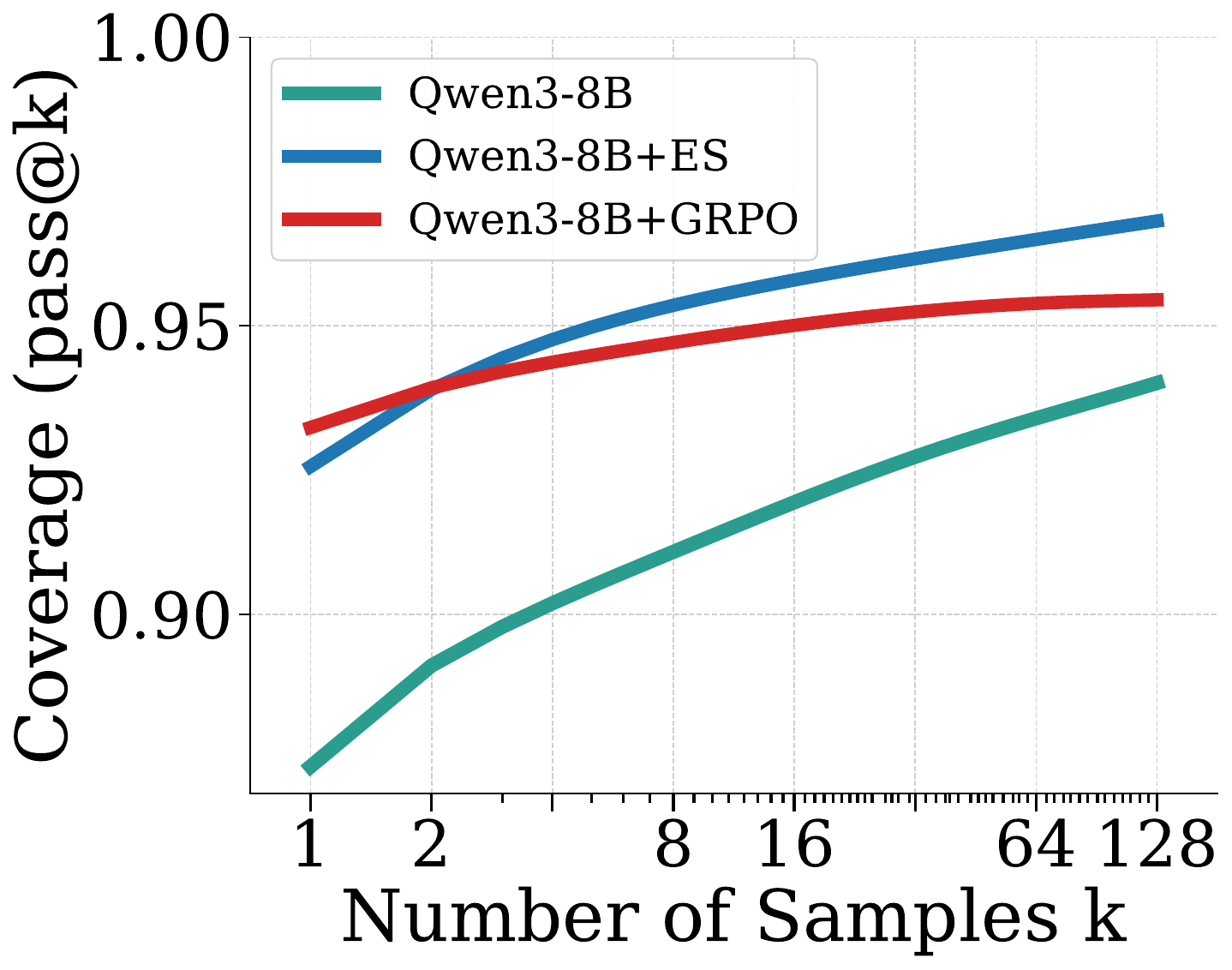}
        \caption{Qwen3-8B}
        \label{fig:qwen3-8b-gsm8k}
    \end{subfigure}
    \begin{subfigure}[b]{0.3\columnwidth}
        \centering
        \includegraphics[width=\textwidth]{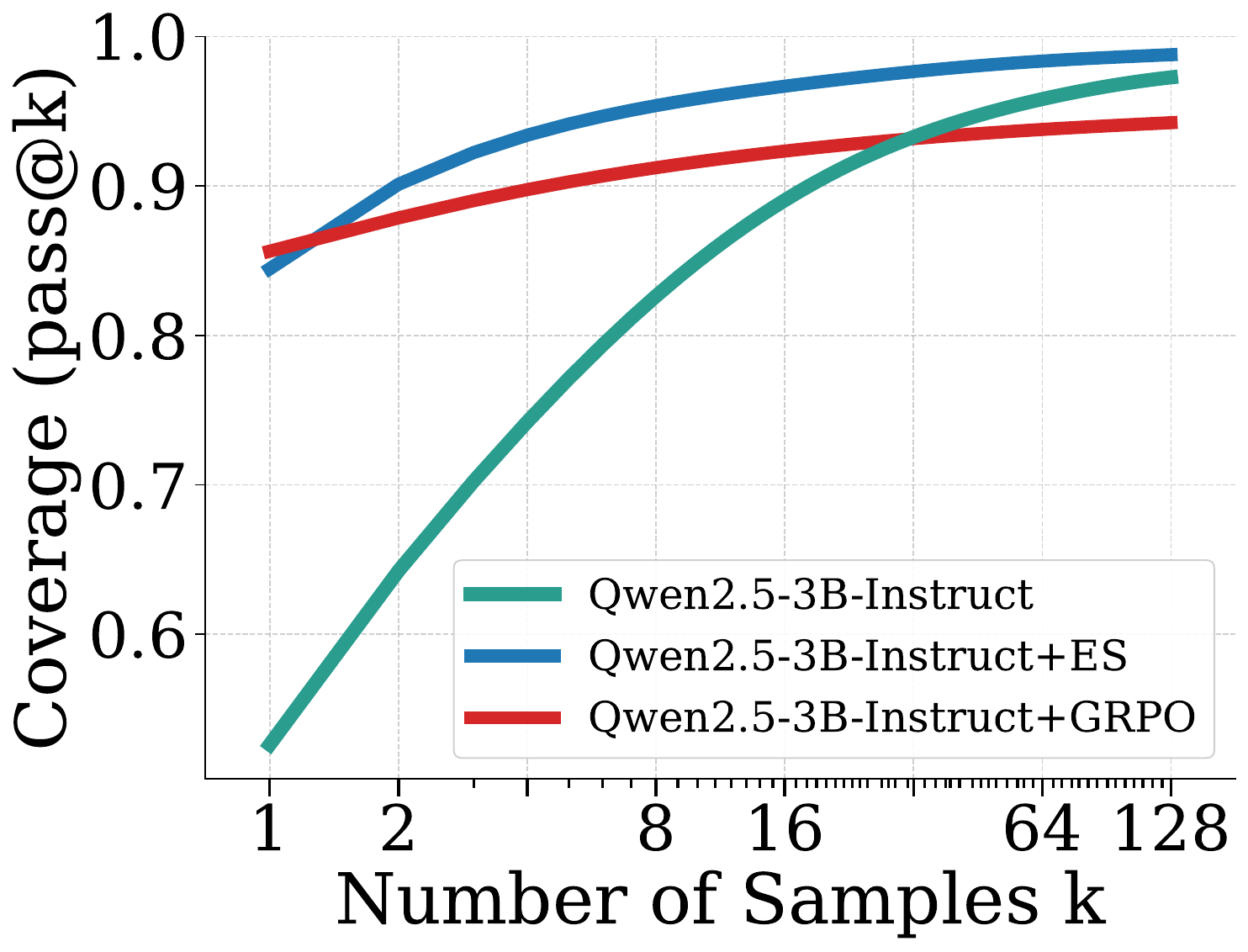}
        \caption{Qwen2.5-1.5B-Instruct}
        \label{fig:qwen2.5-3b-instruct-gsm8k}
    \end{subfigure}
    \hfill
    \begin{subfigure}[b]{0.3\columnwidth}
        \centering
        \includegraphics[width=\textwidth]{plots/qwen2.5-instruct/passk_qwen2_5_3b_instruct_gsm_8k_evalTemp0.6.pdf}
        \caption{Qwen2.5-3B-Instruct}
        \label{fig:qwen2.5-3b-instruct-gsm8k}
    \end{subfigure}
    \hfill
    \begin{subfigure}[b]{0.3\columnwidth}
        \centering
        \includegraphics[width=\textwidth]{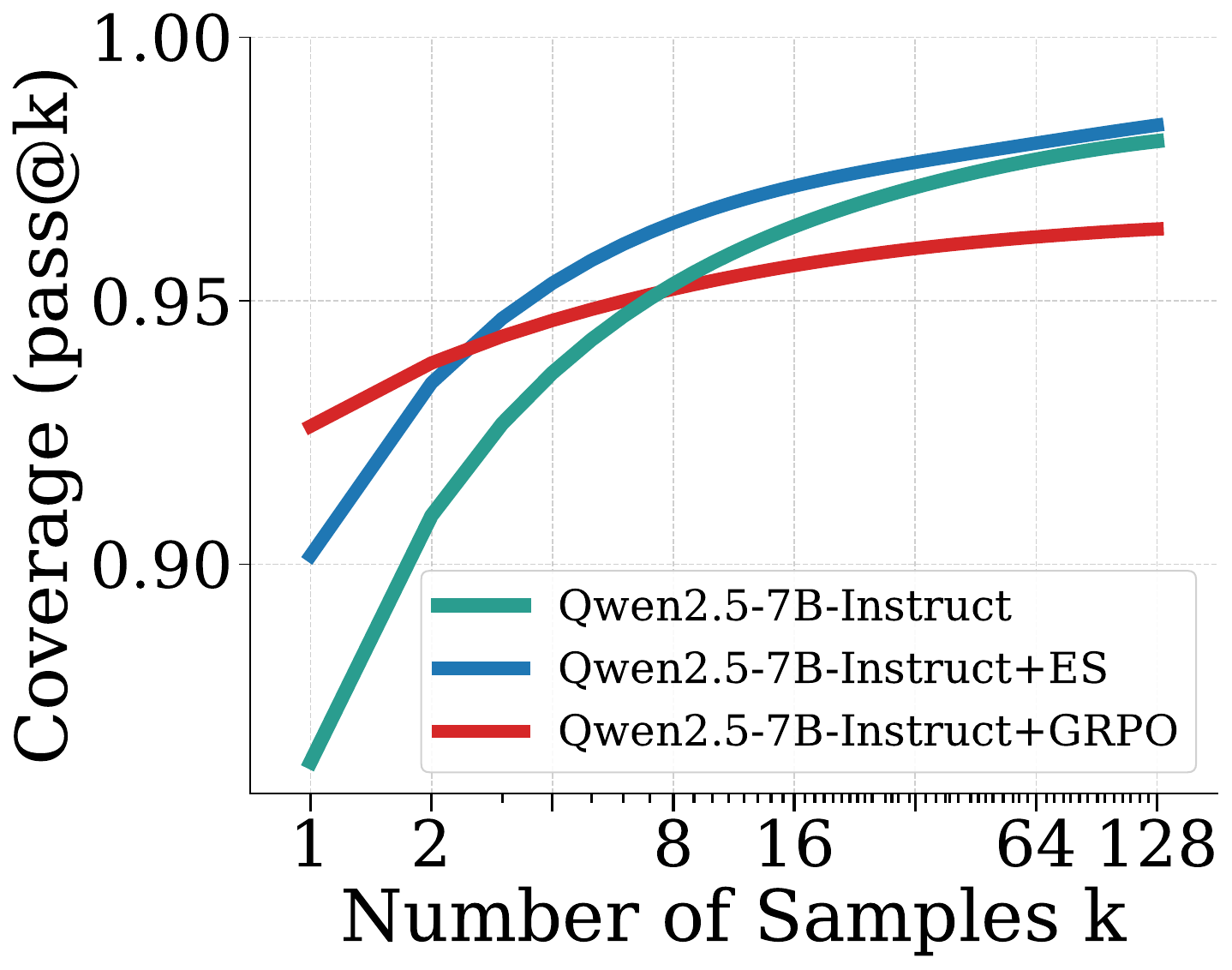}
        \caption{Qwen2.5-7B-Instruct}
        \label{fig:qwen-2.5-7B=gsm8k}
    \end{subfigure}

    \caption{\emph{GSM8K Results.} ES outperforms RL on pass@k for (a--c) Qwen3-1.7B, -4B, -8B and (d--f) Qwen2.5-1.5B, -3B, -7B-Instruct. As $k$ increases, ES continues to improve while RL plateaus and is overtaken by the base model. ES thus improves pass@k without sacrificing pass@1, making it a more suitable post-training method for discovery domains.}
    \label{fig:qwen-gsm8k}
\end{figure}

\paragraph{ES improves pass@k over RL across model families on GSM8K.}
Figure \ref{fig:qwen-gsm8k} shows that ES consistently improves pass@k over RL across Qwen2.5-Instruct and Qwen3 models at scales ranging from 1.5B to 8B. The crossover point, where ES and RL perform equally, typically occurs at $k=2$, after which ES outperforms RL. Notably, the RL pass@k curves plateau earlier than those of ES, consistent with distribution collapse narrowing the solution coverage of RL-trained models. Furthermore, for the Qwen2.5-Instruct models the base model overtakes RL at sufficiently large $k$, empirically confirming the findings of \citet{yue2025does}. The advantage of ES over RL grows with $k$, suggesting that ES becomes increasingly beneficial as the test-time compute budget increases. This pattern is consistent across both model families, supporting the generality of the finding.

\paragraph{ES preserves solution coverage while RL does not.}
Crucially, the phenomenon reported by \citet{yue2025does} that the base model outperforms the RL-fine-tuned model at sufficiently large $k$, is not observed for ES. This suggests that ES post-training improves pass@1 without sacrificing the broad output distribution support of the base model. As a result, ES models benefit from increasing test-time compute in a way that RL models do not, making ES a potentially more suitable post-training strategy for discovery problems (see Section \ref{sec:test_time_scaling}).

\subsection{MATH Results}
\label{subsec:math_experiments}

\begin{figure*}
    \centering

    \begin{subfigure}[b]{0.3\columnwidth}
        \centering
        \includegraphics[width=\textwidth]{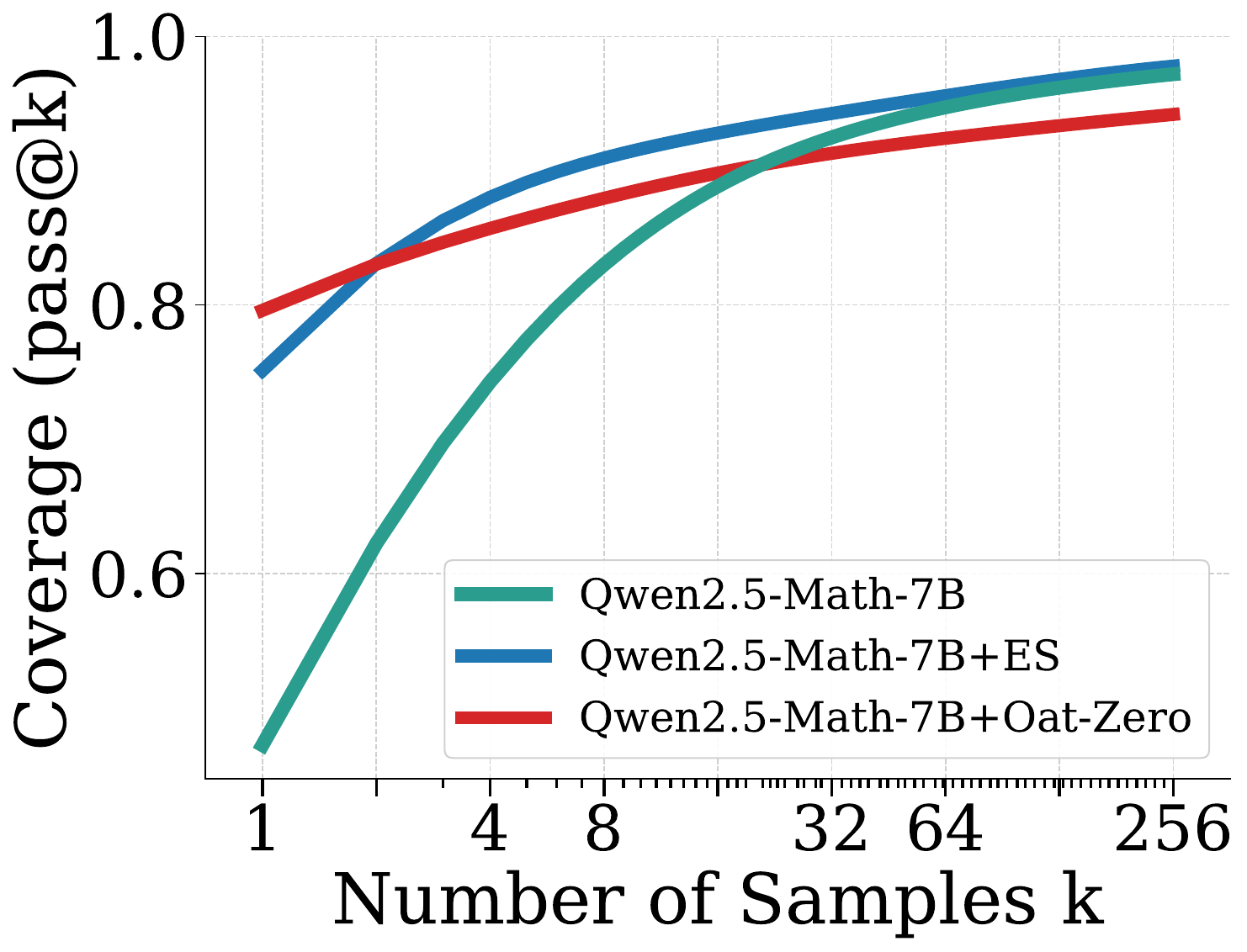}
        \caption{MATH500}
        \label{fig:fig1}
    \end{subfigure}
    %\hspace{20pt}
    \hfill
    \begin{subfigure}[b]{0.3\columnwidth}
        \centering
        \includegraphics[width=\textwidth]{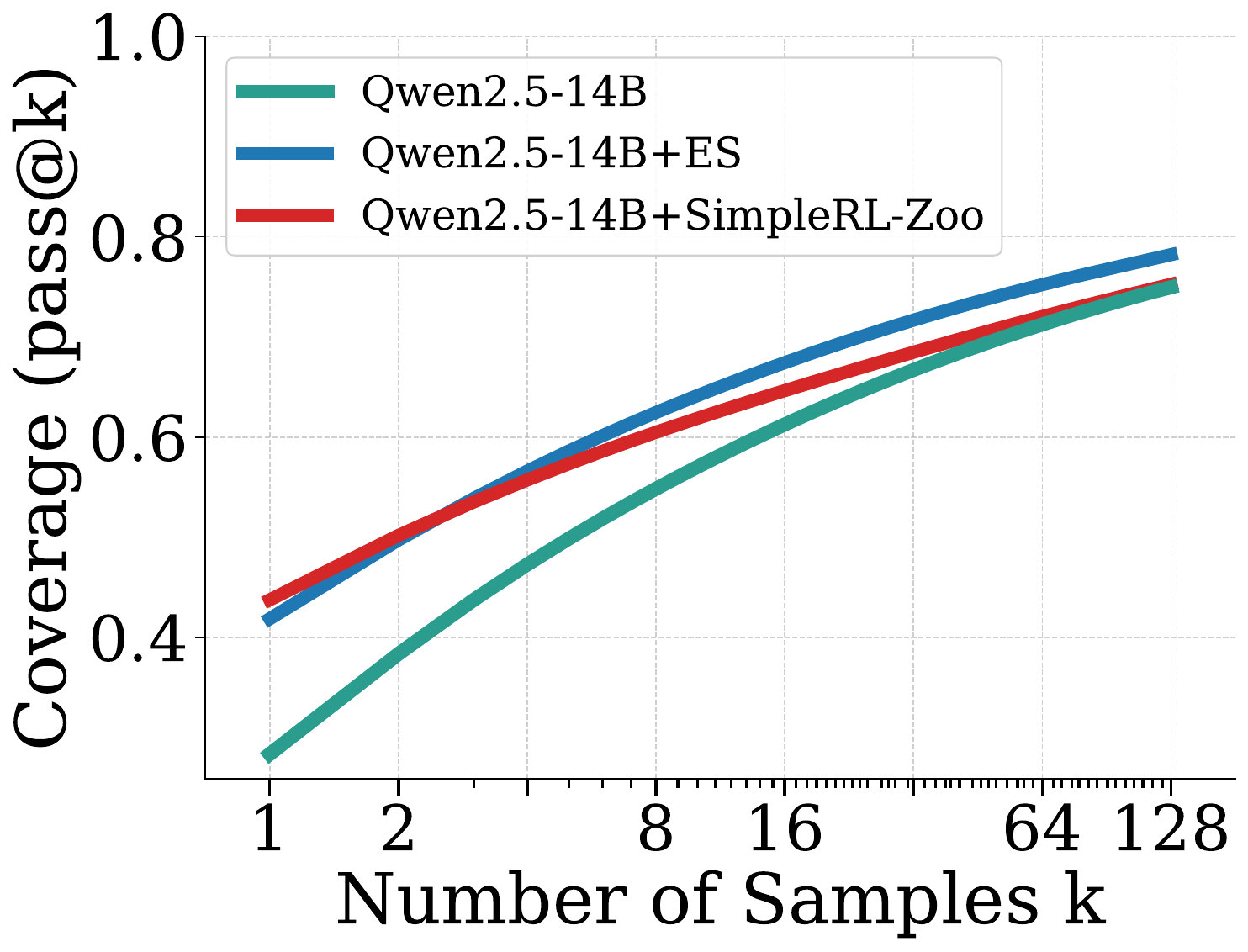}
        \caption{Olympiad Bench}
        \label{fig:fig2}
    \end{subfigure}
    %\hspace{20pt}
    \hfill
    \begin{subfigure}[b]{0.3\columnwidth}
        \centering
        \includegraphics[width=\textwidth]{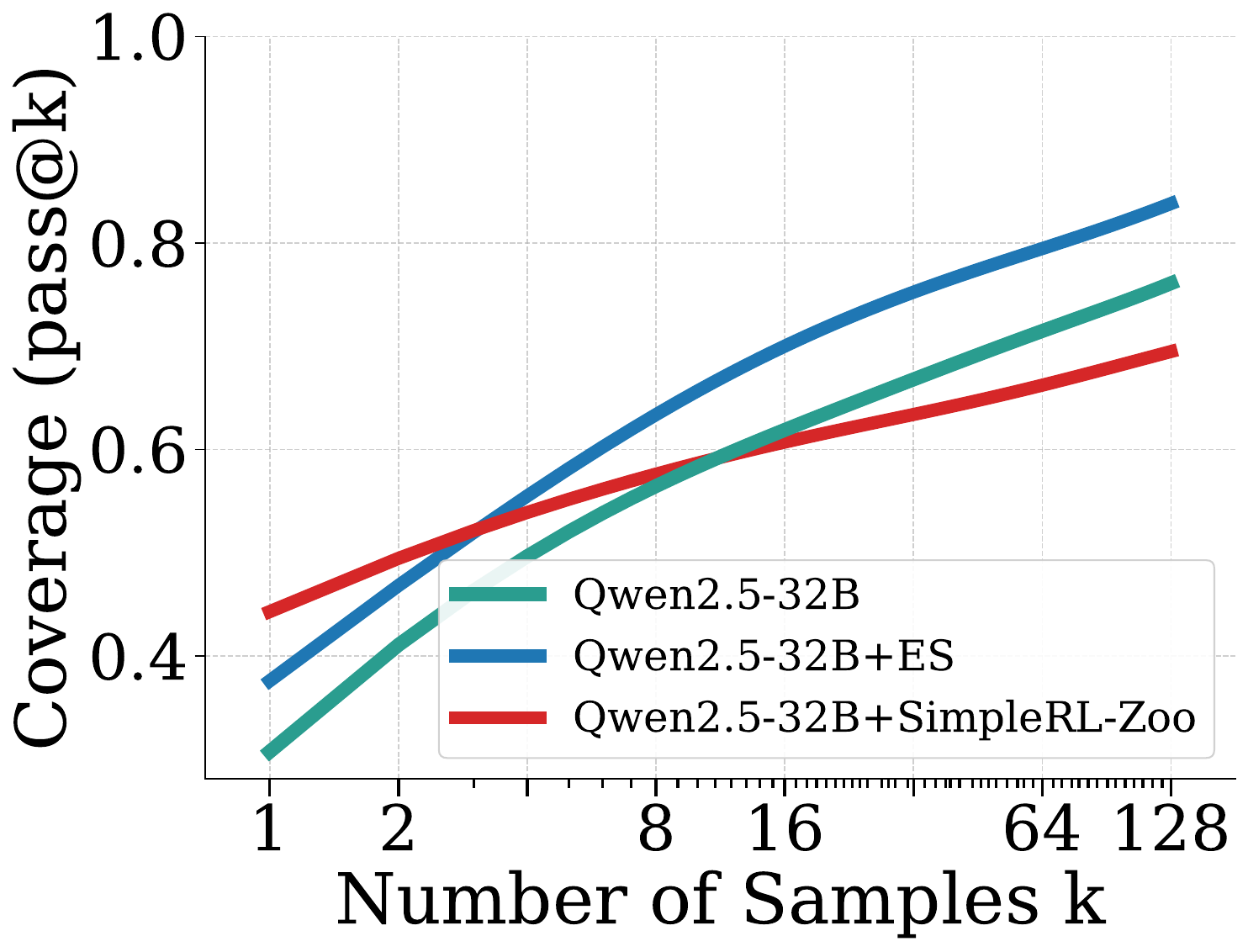}
        \caption{Minerva}
        \label{fig:qwen2.5-32b-minerva}
    \end{subfigure}

    \caption{\emph{Comparison to state-of-the-art RL checkpoints.} ES outperforms SOTA RL checkpoints (OatZero and SimpleRL-Zoo) on pass@k, across (a) 7B, (b) 14B, and (c) 32B parameter models, showing that the advantage of ES persists as models scale. This result further motivates the use of ES in TTS scenarios with larger models.
    See Appendix \ref{appendix} Figure~\ref{fig:appendix-qwen2.5-passk} for additional results.}
    \label{fig:qwen2.5-32b}
\end{figure*}

Next, pass@k for ES and RL was compared by evaluating Qwen2.5-Math-7B, Qwen2.5-14B, and Qwen2.5-32B on MATH500, Olympiad Bench, and Minerva benchmarks. 

\paragraph{ES improves pass@k over RL across math benchmarks.}
Figure \ref{fig:qwen2.5-32b} shows that ES consistently improves pass@k over RL across Qwen2.5-Math-7B, Qwen2.5-14B, and Qwen2.5-32B. %on MATH500, Olympiad Bench, and Minerva.
For Qwen2.5-Math-7B, ES outperforms OatZero for $k > 1$ across on MATH500. For Qwen2.5-14B, ES outperforms SimpleRL-Zoo for $k > 1$ on Olympiad Bench. Notably, for Qwen2.5-14B the base model becomes competitive with SimpleRL-Zoo at large $k$ on Olympiad Bench.
For Qwen2.5-32B, ES retains a clear advantage over SimpleRL-Zoo on Minerva where the gap grows with $k$.
Results for additional model/benchmark combinations are presented in Figure \ref{fig:appendix-qwen2.5-passk} Appendix \ref{appendix} and highlight that ES outperforms RL across larger models for each measured benchmark.

\paragraph{ES preserves solution coverage while RL does not.}
The MATH experiments provided further evidence that distribution collapse is not an artifact of a specific RL algorithm but a general consequence of RL post-training. The RL-trained models exhibit earlier pass@k saturation compared to ES, especially on Minerva, consistent with a narrowing of the output distribution support. Crucially, the base model does not outperform the ES fine-tuned model at any value of $k$ or model scale tested, showing that ES post-training improves pass@1 without sacrificing the broad output distribution support of the base model. This pattern holds consistently from 7B to 32B parameters, indicating that ES's ability to preserve solution coverage is not a small-scale artifact but persists as models grow substantially larger.

\begin{figure}[t!]
    \centering

    \begin{subfigure}[b]{0.4\columnwidth}
        \centering
        \includegraphics[width=\textwidth]{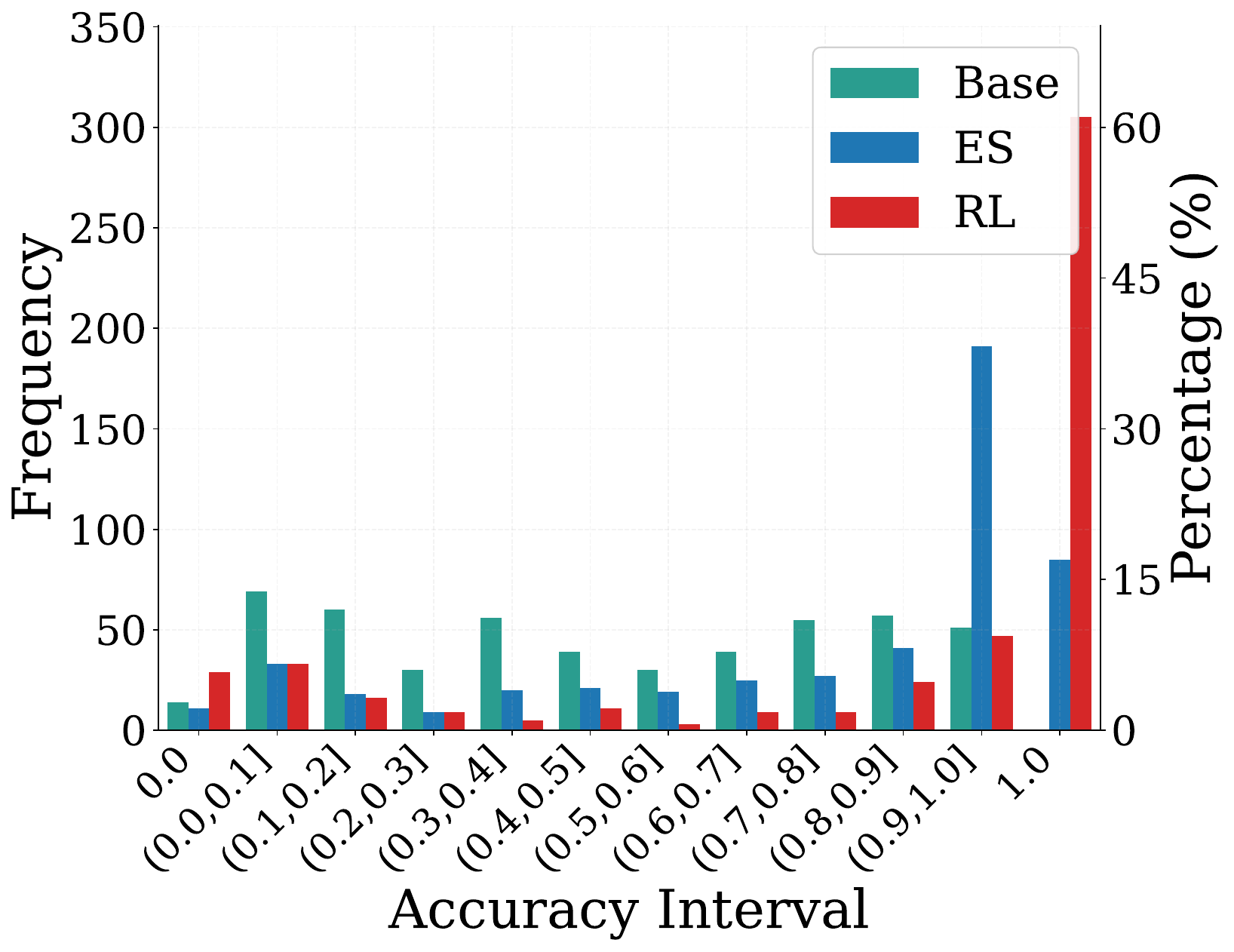}
        \caption{MATH500}
        \label{fig:accuracy-dist-qwen2.5-math-7b-math500}
    \end{subfigure}
    \hspace{0.03\textwidth}
    \begin{subfigure}[b]{0.4\columnwidth}
        \centering
        \includegraphics[width=\textwidth]{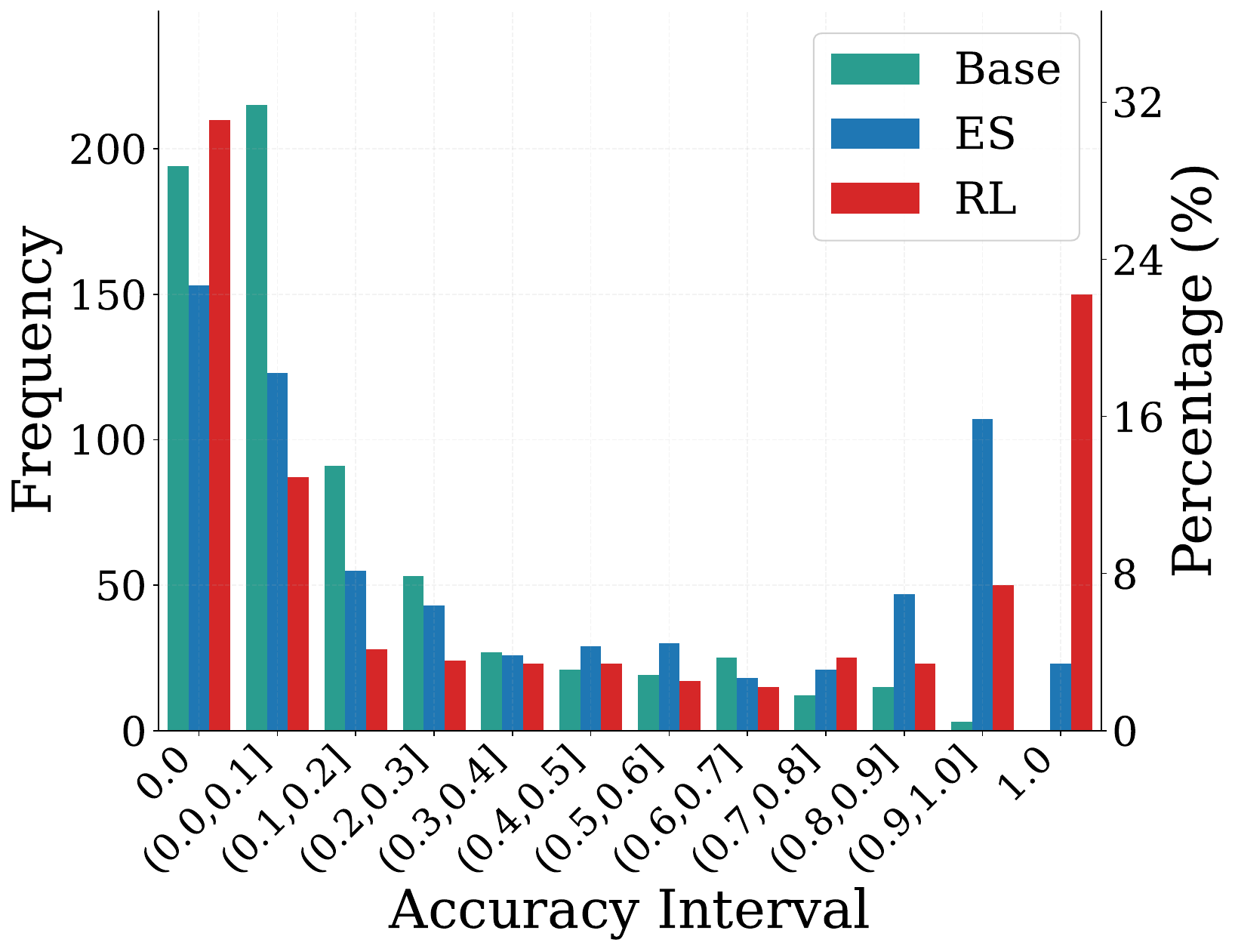}
        \caption{Olympiad Bench}
        \label{fig:accuracy-dist-qwen2.5-math-7b-olympiadbench}
    \end{subfigure}       

    \caption{\emph{Qwen2.5-Math-7B accuracy distributions for ES and RL (OatZero) on MATH500 and Olympiad Bench.} Bin 0.0 measures unsolvable problems, i.e., where none of the model's responses are correct. Relative to the base model, RL increases unsolvable problems while ES reduces them.}
    \label{fig:accuracy-dist}
\end{figure}

\section{Solution Coverage}
\label{sec:solution-coverage}

To better understand \emph{how} ES achieves such high pass@k compared to RL or the base model, this section analyzes model behavior from three perspectives: accuracy distributions, regressions vs. progressions, and answer entropy.

\paragraph{Accuracy distribution analysis.}
Figure \ref{fig:accuracy-dist} shows the accuracy distribution for ES and RL Qwen2.5-Math-7B models on  MATH500 and Olympiad Bench. Each bin shows the mean accuracy across $k$ responses for a given prompt, where the frequency on the y-axis denotes the number of prompts in each bin over the full dataset.
For example, bin $1.0$ contains prompts where all $k$ responses were correct, and bin $0.0$ contains prompts where all $k$ responses were incorrect.

ES and RL affect the accuracy distribution differently with respect to the base model.
RL increases the frequency in bin $1.0$ relative to the base model, reflecting its objective of maximizing pass@1.
ES also increases frequency in bin $1.0$, but the improvement is more distributed across bin $(0.9, 1.0]$, consistent with the pass@1 differences between ES and RL reported in Section \ref{sec:pass@k-math}.

A notable artifact of RL training is an increase in the frequency of bin $0.0$ relative to the base model, which was also observed by \citet{yue2025does}.
This indicates that RL renders a subset of prompts that the base model could solve entirely unsolvable across $k$ samples, reducing solution coverage and limiting the model's ability to generalize across a broad range of problems. RL's increase in unsolvable prompts imposes a hard ceiling on achievable pass@k performance that no amount of additional sampling can overcome.
In contrast, ES reduces the frequency in bin $0.0$ relative to the base model, improving solution coverage across all benchmarks. Figures \ref{fig:appendix-gsm8k-accuracy-distribution} and \ref{fig:appendix-math-accuracy-distribution} in Appendix \ref{appendix} show the accuracy distributions for Qwen2.5, Qwen3, and Qwen2.5-Math across scales from $1.5$B to $32$B, consistently showing that ES broadens solution coverage while RL narrows it.

\paragraph{Measuring model regressions and progressions.}
The narrowing and broadening of solution coverage can be quantified through model \emph{progressions} and \emph{regressions}. Progressions measure knowledge gained during fine-tuning: prompts the base model got wrong but the fine-tuned model answered correctly.
Regressions measure knowledge lost during fine-tuning: prompts the base model answered correctly but the fine-tuned model got wrong.
Ideally, a post-trained model maximizes progressions and minimizes regressions.

\begin{figure}
    \centering
    \hfill
    \begin{subfigure}[b]{0.3\columnwidth}
        \centering
        \includegraphics[width=\textwidth]{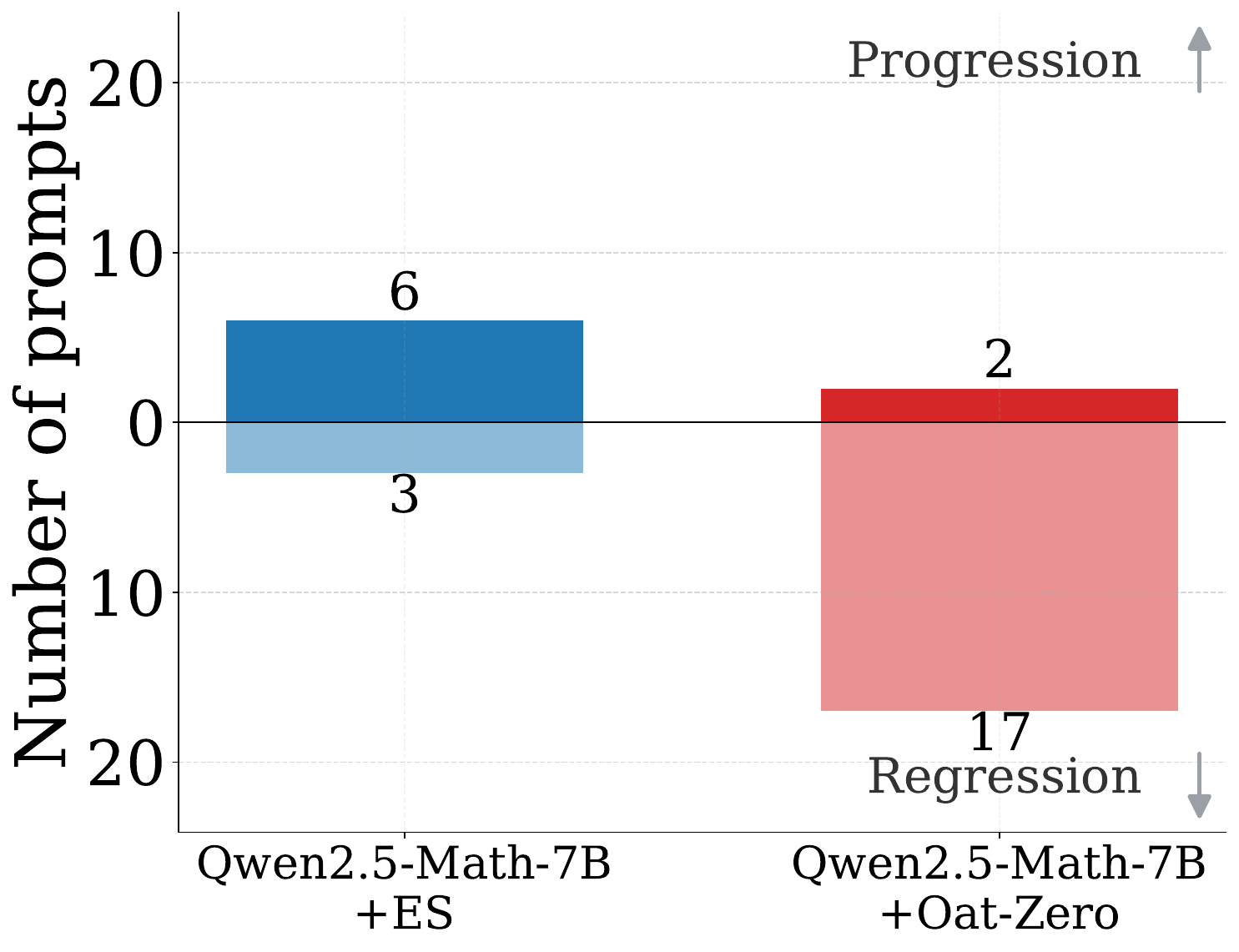}
        \caption{MATH500}
        \label{fig:prog-reg-qwen2.5-math-7b-math500}
    \end{subfigure}
    \hfill
    \begin{subfigure}[b]{0.3\columnwidth}
        \centering
        \includegraphics[width=\textwidth]{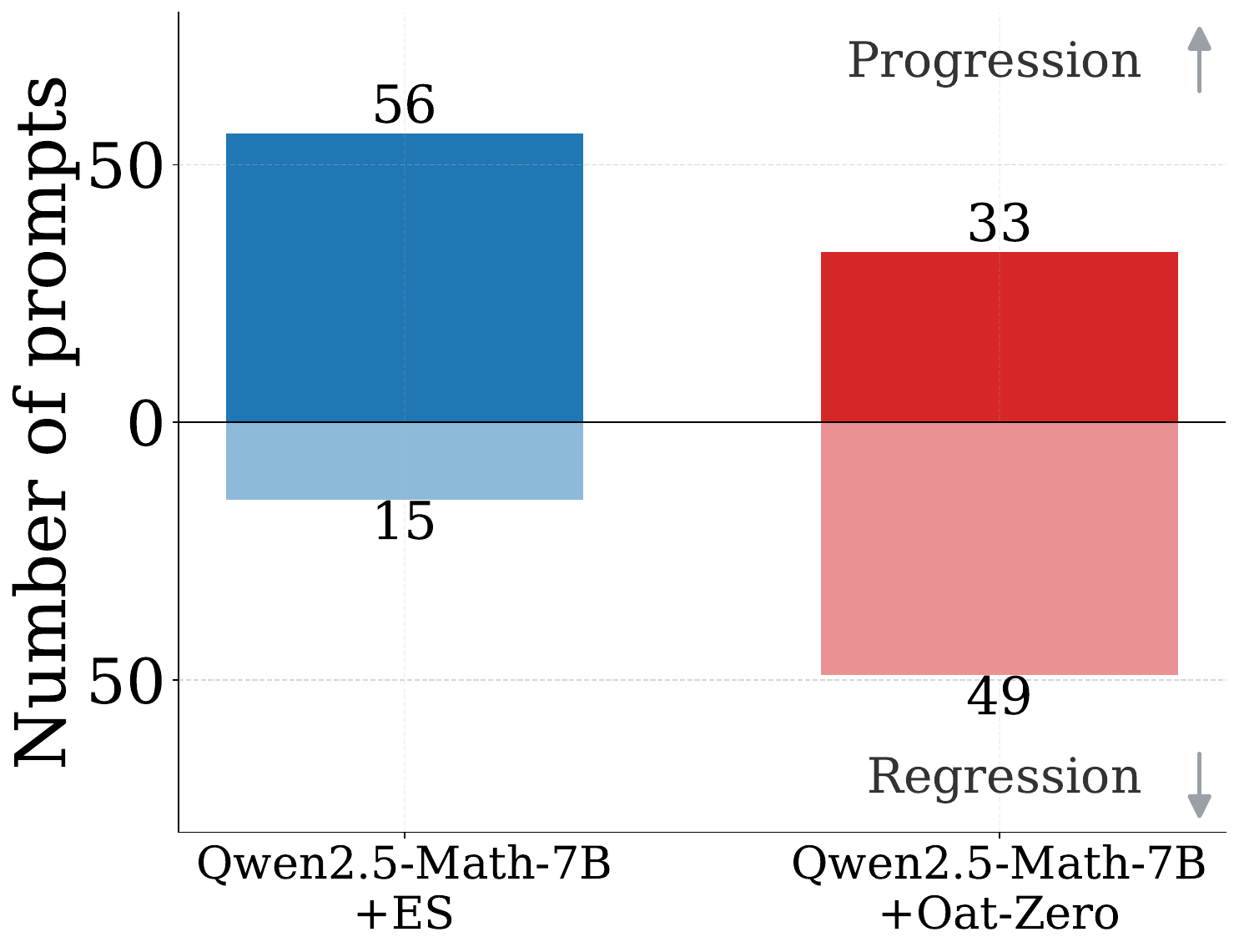}
        \caption{Olympiad Bench}
        \label{fig:prog-reg-qwen2.5-math-7b-olympiad-bench}
    \end{subfigure}
    \hfill
    \begin{subfigure}[b]{0.3\columnwidth}
        \centering
        \includegraphics[width=\textwidth]{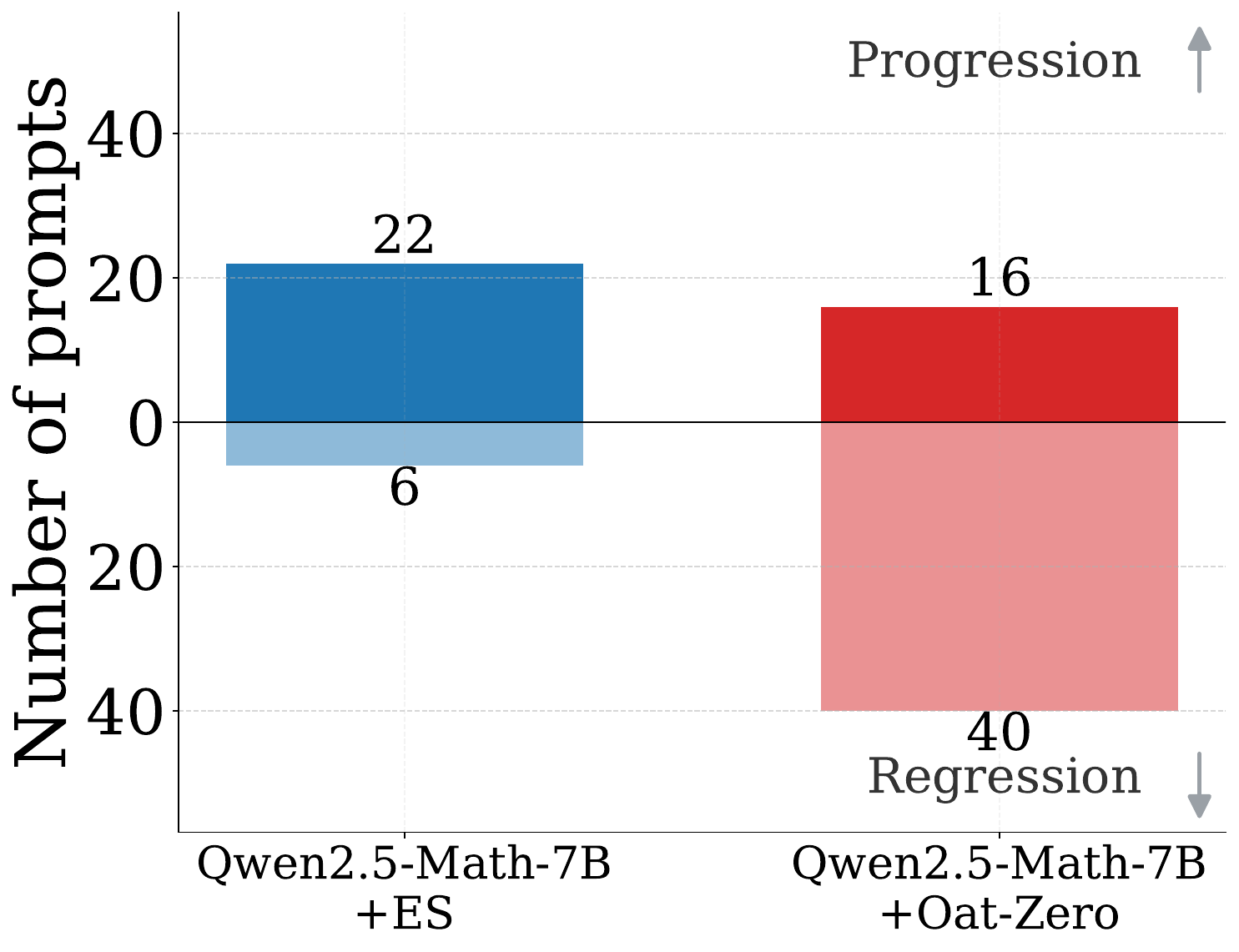}
        \caption{Minerva}
        \label{fig:prog-reg-qwen2.5-math-7b-minerva}
    \end{subfigure}
    \hfill
    \caption{\emph{Progressions and regressions for Qwen2.5-Math-7B evaluated using MATH500, Olympiad Bench, and Minerva.} ES reduces regressions and increases progressions across all three benchmarks compared to RL fine-tuned model checkpoints (\emph{OatZero}).}
    \label{fig:prog-reg}
\end{figure}

Figures \ref{fig:prog-reg-qwen2.5-math-7b-math500}, \ref{fig:prog-reg-qwen2.5-math-7b-olympiad-bench}, and \ref{fig:prog-reg-qwen2.5-math-7b-minerva} report progressions and regressions for ES and RL fine-tuned Qwen2.5-Math-7B models.
ES yields significant increases of progressions compared to RL: There are $6$ for ES vs. $3$ for RL on MATH500; $56$ vs. $33$ on Olympiad Bench; and $22$ vs. $16$ on Minerva.
A natural concern when comparing ES to RL is that ES may preserve the base model distribution simply because it is not learning effectively.
The progression results show ES adds more new correct solutions than RL, i.e.\ ES improves the model's knowledge rather than simply not damaging it. 
ES also results in significantly fewer regressions compared to RL. For Qwen2.5-Math-7B, ES has $3$ regressions vs. $17$ for RL on MATH500; on Olympiad Bench, ES has $15$ vs. $49$ for RL; on Minerva, ES has $6$ vs. $40$ for RL. Additional progression and regression results are included in Figures \ref{fig:appendix-progressions-regressions-gsm8k} and \ref{fig:appendix-progressions-regressions-math} Appendix \ref{appendix}, showing that across Qwen2.5 and Qwen3 models from 1.5B to 32B parameter scale, ES increases progressions and reduces regressions over RL.

Regressions are a signal of catastrophic forgetting, where knowledge already contained within the base model is lost. RL overwrites some of this knowledge in order to maximize reward on the training distribution, whereas ES pushes parameters toward flat regions of weight space and is therefore less destructive to existing knowledge. 
Notably, the gains in progressions from ES are observed \emph{alongside} reductions in regressions, meaning ES is not trading one for the other but improving on both simultaneously. This finding suggests ES fine-tuned models are more quality-preserving with respect to the output distribution and may generalize better to out-of-distribution problems.

Together, the progression and regression results reveal that ES achieves a more favorable knowledge trade-off than RL during fine-tuning. RL gains new knowledge at the cost of overwriting existing knowledge, while ES adds new knowledge while preserving more of what the base model already knew. In other words, ES finds regions of parameter space that are both more capable and more knowledge-preserving. Critically, this dual advantage directly explains the pass@k results, where fewer regressions broaden the lower end of the solution coverage distribution, while more progressions expand the upper end, together producing the consistently higher pass@k curves observed for ES across all benchmarks and model scales. These results thus further strengthen the case for ES as a powerful post-training method.

\begin{figure}[t!]
    \centering

    \begin{subfigure}[b]{0.4\columnwidth}
        \centering
        \includegraphics[width=\textwidth]{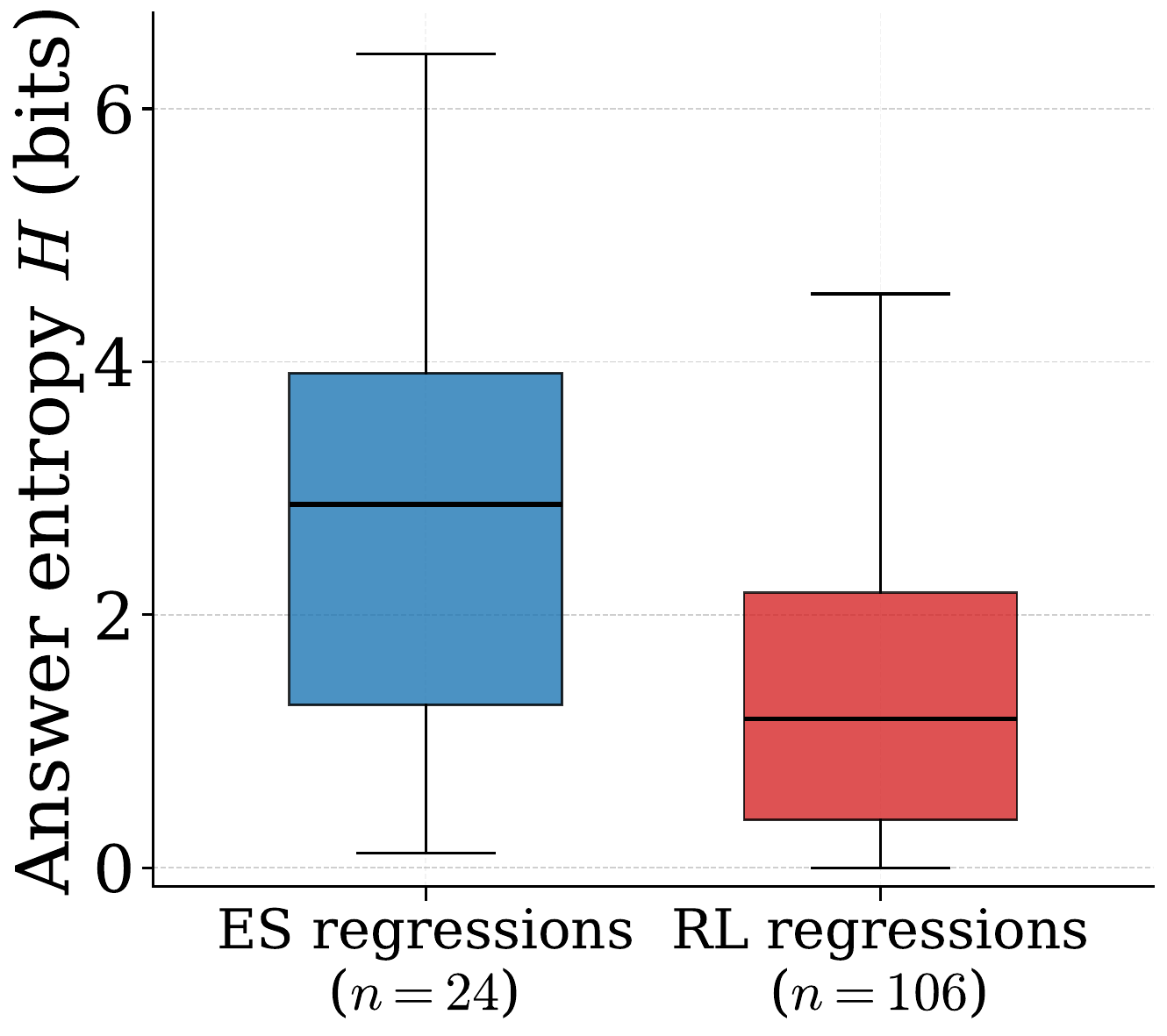}
        \caption{}
        \label{fig:entropy-regressions-qwen2.5-math-7b-box}
    \end{subfigure}
    \hspace{0.03\textwidth}
    \begin{subfigure}[b]{0.4\columnwidth}
        \centering
        \includegraphics[width=\textwidth]{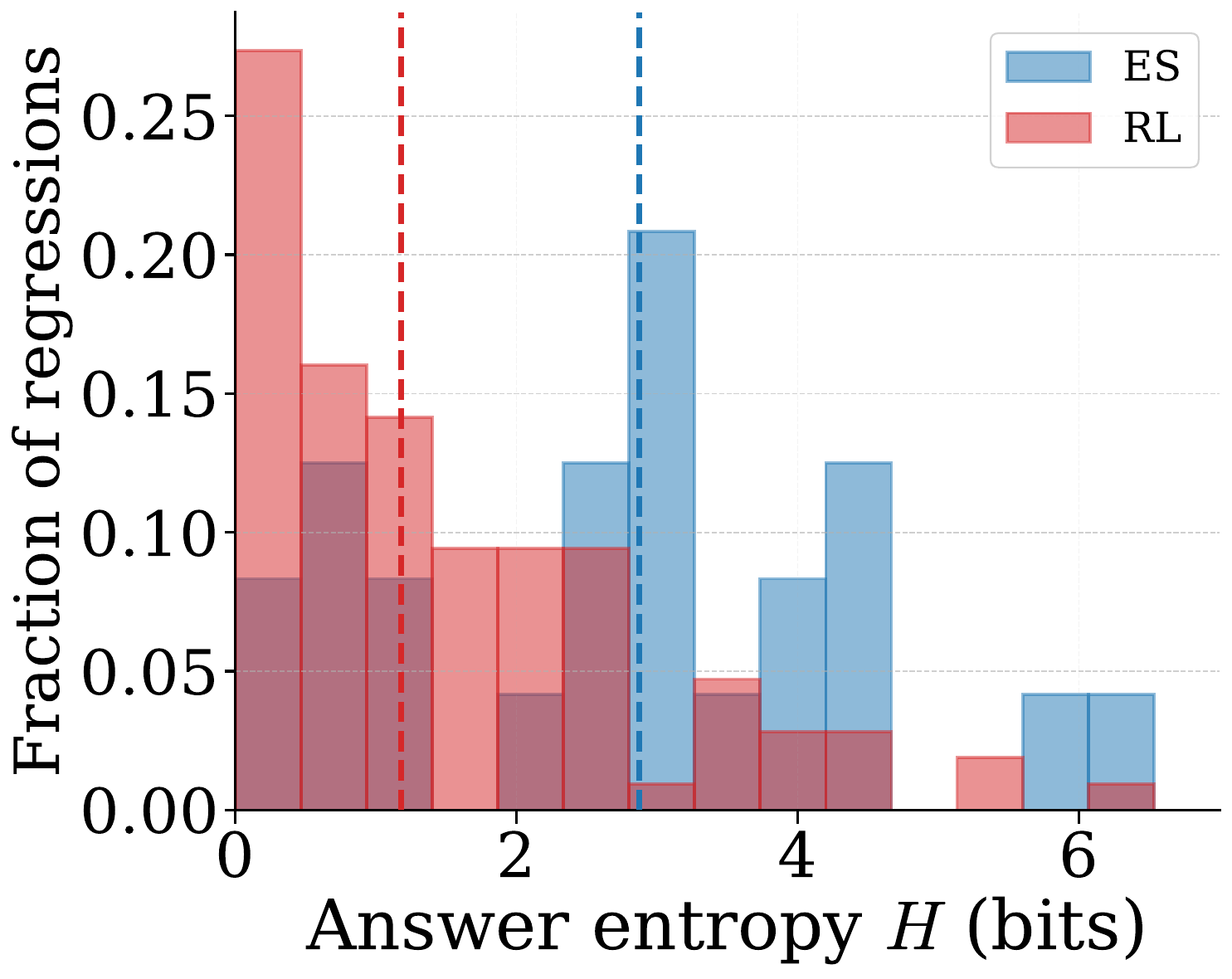}
        \caption{}
        \label{fig:entropy-regressions-qwen2.5-math-7b-dist}
    \end{subfigure}

    \caption{\emph{Entropy of the answer distribution plots of combined model regressions for ES and RL across MATH500, Olympiad Bench, and Minerva for Qwen2.5-Math-7B.} (a) ES maintains higher entropy when compared to RL during failures. (b) ES maintains a distribution over regressions at a higher entropy, while RL maintains low entropy during regressions.}
    \label{fig:entropy-regressions-qwen2.5-math-7b}
\end{figure}

\paragraph{Entropy analysis for ES and RL failure modes.}
Figure \ref{fig:entropy-regressions-qwen2.5-math-7b} shows the entropy over answer distributions for ES and RL regressions for Qwen2.5-Math-7B. The answer distribution is constructed by taking the unique final answers per problem and computing a frequency distribution, over which the Shannon entropy is calculated \citep{shannon1948mathematical}. ES has a total of $24$ regressions across MATH500, Olympiad Bench, and Minerva, while RL has $106$. Figure \ref{fig:entropy-regressions-qwen2.5-math-7b-box} shows that ES maintains higher entropy than RL, indicating that when ES fails it preserves greater diversity over candidate answers.
Figure \ref{fig:entropy-regressions-qwen2.5-math-7b-dist} presents the distribution over entropy values, where each bin corresponds to the fraction of total regressions across all three benchmarks. Notably, more than $25\%$ of RL regressions have entropy close to zero, meaning the model collapses to a small number of responses and is confidently incorrect.
In contrast, the ES entropy distribution is shifted to the right relative to RL, with a higher median entropy, further confirming that ES failures are characterized by uncertainty rather than confident convergence to incorrect answers.

\begin{figure}
    \centering

    \begin{subfigure}[b]{0.4\columnwidth}
        \centering
        \includegraphics[width=\textwidth]{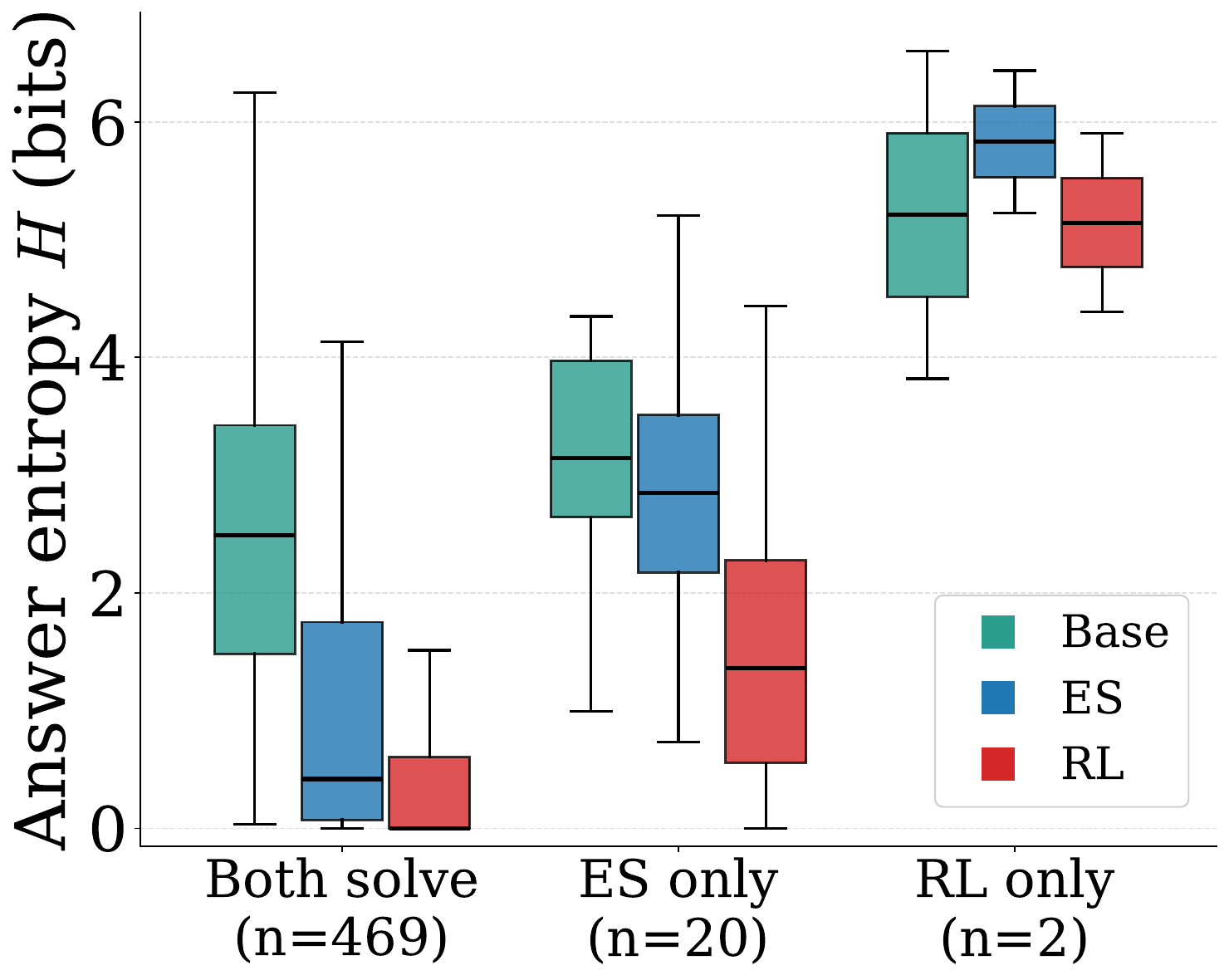}
        \caption{MATH500}
        \label{fig:prog-reg-qwen3-1.7b-gsm8k}
    \end{subfigure}
    \hspace{0.03\textwidth}
    \begin{subfigure}[b]{0.4\columnwidth}
        \centering
        \includegraphics[width=\textwidth]{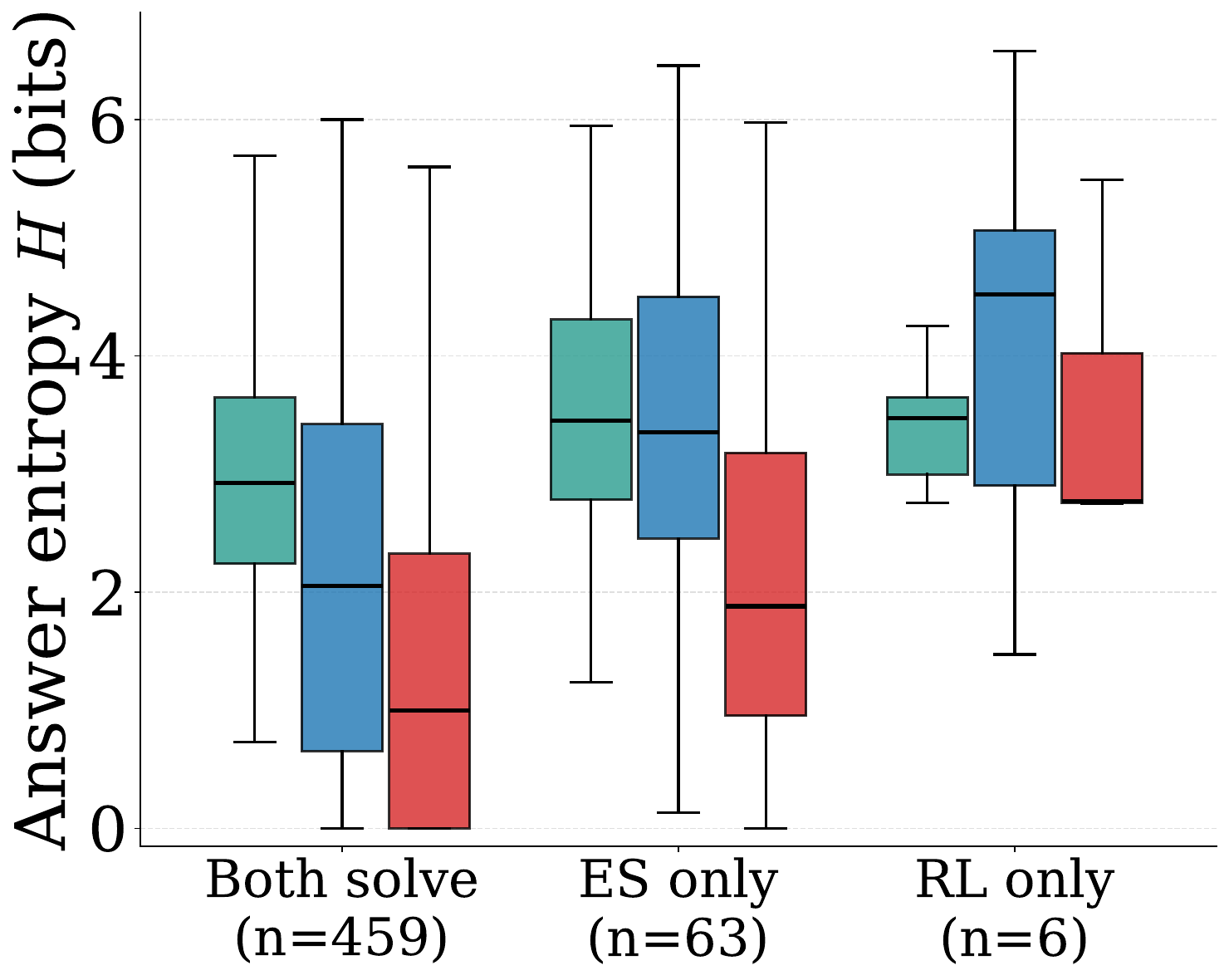}
        \caption{Olympiad Bench}
        \label{fig:prog-reg-qwen3-4b-gsm8k}
    \end{subfigure}

    \caption{\emph{Box plots of answer entropy for Qwen2.5-Math-7B on MATH500 and Olympiad Bench.} The box plots are clustered into subsets where both ES and RL fail on specific problems. For each set, RL fails confidently and narrowly (low entropy), whereas ES fails with preserved uncertainty and succeeds with reasonable confidence (high entropy).}
    \label{fig:box-plots-qwen2.5-math-7b-failure-modes}
\end{figure}

\begin{figure}[t!]
    \centering
    % Row 1: a, b, c
    % \begin{subfigure}[b]{0.225\textwidth}
    %     \centering
    %     \includegraphics[width=\textwidth]{plots/qwen2.5-math-7b/prog_reg_qwen2_5_math_7b_math500_evalTemp0.6.pdf}
    %     \caption{MATH500}
    %     \label{fig:prog-reg-qwen2.5-math-7b-math500}
    % \end{subfigure}
    % \hfill
        \begin{subfigure}[b]{0.31\columnwidth}
        \centering
        \includegraphics[width=\textwidth]{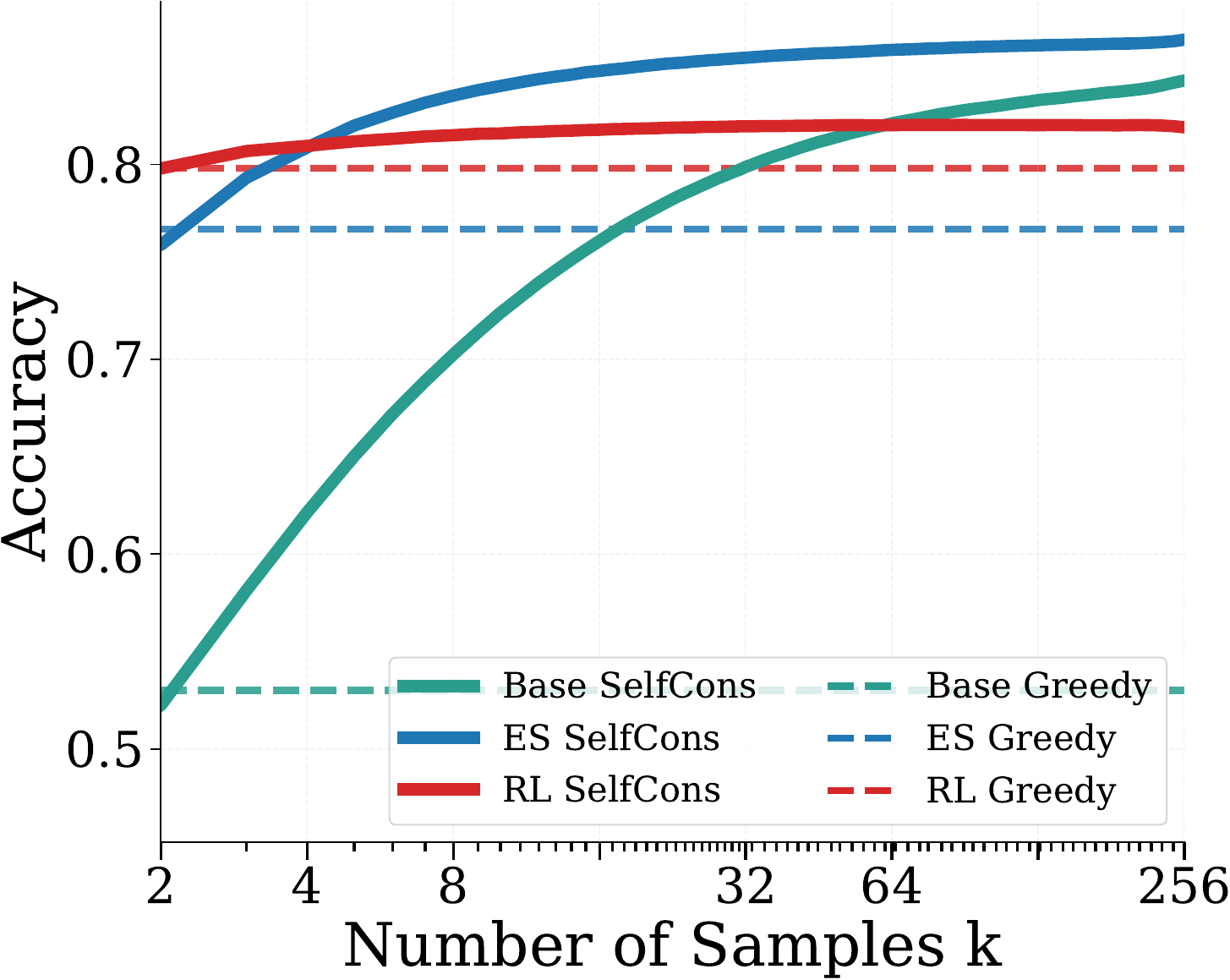}
        \caption{MATH500}
        \label{fig:self_cons_math500}
    \end{subfigure}
    \hfill
    %     \begin{subfigure}[b]{0.225\textwidth}
    %     \centering
    %     \includegraphics[width=\textwidth]{plots/qwen2.5-math-7b/prog_reg_qwen2_5_math_7b_olympiad_bench_evalTemp0.6.pdf}
    %     \caption{Olympiad Bench}
    %     \label{fig:prog-reg-qwen2.5-math-7b-olympiad-bench}
    % \end{subfigure}
    % \hfill
    \begin{subfigure}[b]{0.31\columnwidth}
        \centering
        \includegraphics[width=\textwidth]{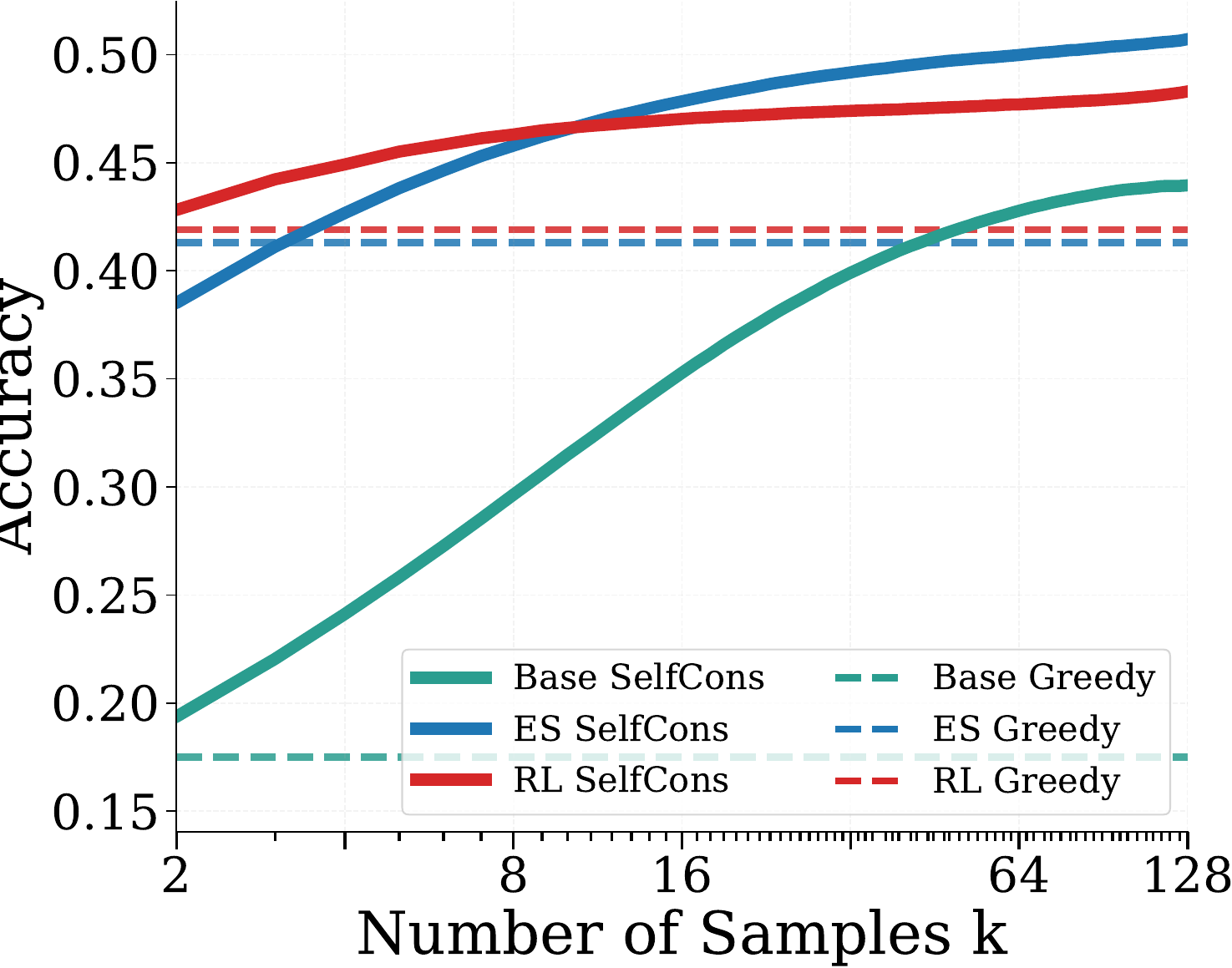}
        \caption{Olympiad Bench}
        \label{fig:self_cons_olympiad}
    \end{subfigure}
    \hfill
    % \begin{subfigure}[b]{0.22\textwidth}
    %     \centering
    %     \includegraphics[width=\textwidth]{plots/qwen2.5-math-7b/prog_reg_qwen2_5_math_7b_minerva_evalTemp0.6.pdf}
    %     \caption{Minerva}
    %     \label{fig:prog-reg-qwen2.5-math-7b-minerva}
    % \end{subfigure}
    % \hfill
    \begin{subfigure}[b]{0.31\columnwidth}
        \centering
        \includegraphics[width=\textwidth]{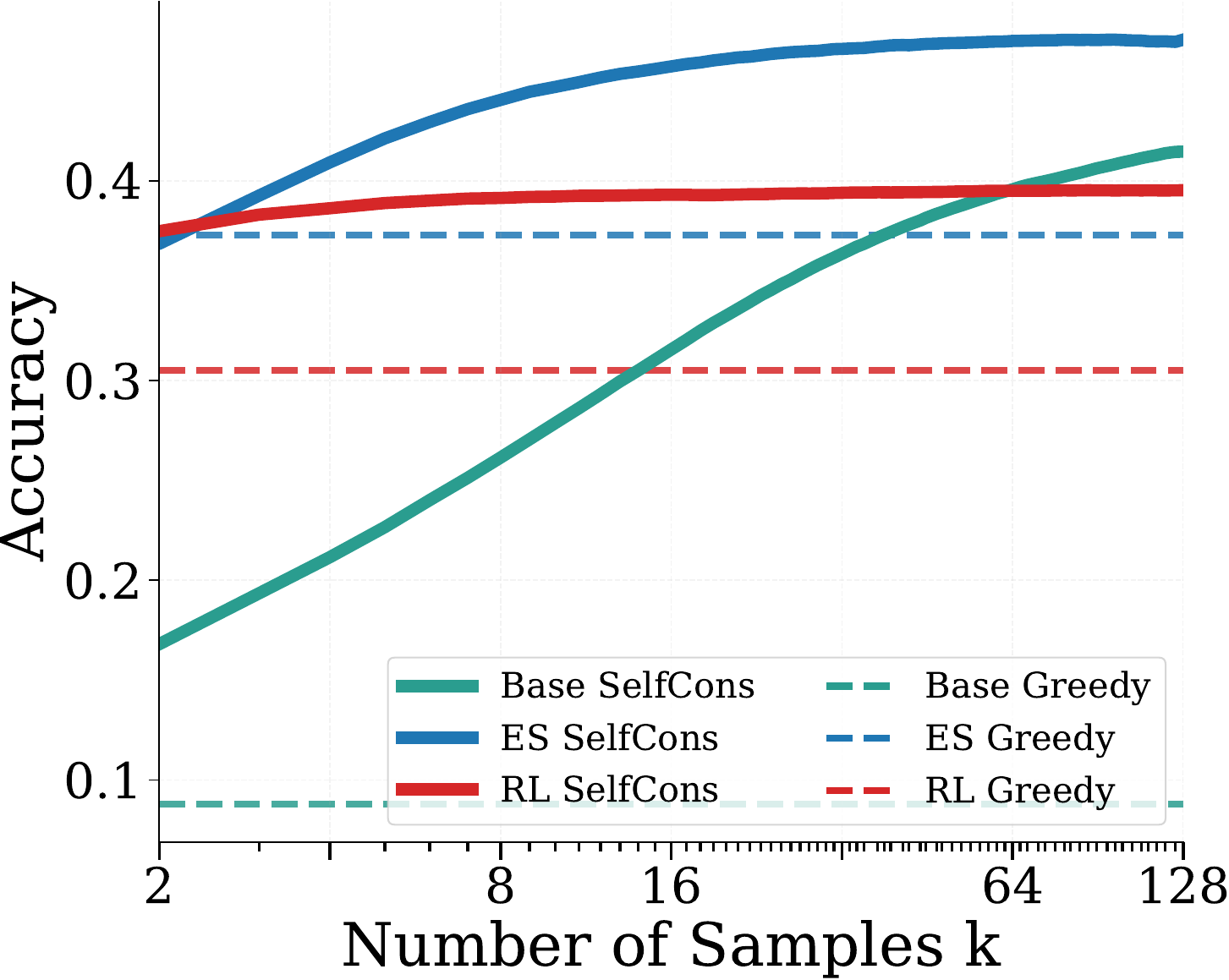}
        \caption{Minerva}
        \label{fig:self_cons_minerva}
    \end{subfigure}

    \caption{\emph{ES and RL performance measuring self-consistency (voting)}. ES's higher-quality distributions yield higher voting accuracy than RL or the base model as $k$ increases. Each curve averages $1$K random permutations of answers from Section~\ref{subsec:math_experiments}.}
    \label{fig:self-consistency}
\end{figure}

Figure \ref{fig:box-plots-qwen2.5-math-7b-failure-modes} provides a complementary view, presenting box plots of the entropy over the answer distribution per prompt across three subsets of problems: those that both ES and RL solve, those that only ES solves, and those that only RL solves. On problems that both methods solve, ES and RL reduce entropy relative to the base model, indicating increased confidence on solvable problems. A striking difference emerges on problems where methods fail: on problems that ES solves but RL fails, RL exhibits low entropy, indicating that the model confidently converges to a small subset of incorrect answers. In contrast, on problems that RL solves but ES fails, ES preserves or exceeds the entropy of the base model, maintaining solution diversity even in its failure modes. This asymmetry reveals a fundamental difference in how ES and RL fail: whereas RL fails confidently and narrowly, ES fails with preserved uncertainty, a property that is more amenable to correction through increased test-time compute.

\section{Test-time Scaling Beyond Pass@k: Voting} %through Self-consistency}
\label{sec:test_time_scaling}

Beyond its role as a tool for understanding distribution quality, pass@k can be put to good use through TTS. With high-quality output distributions, better solutions are more likely to surface over increased inference steps.
However, pass@k itself is a practical TTS method only in verifiable test settings, e.g., in coding when there are a given set of tests that can be used to programatically judge whether a particular solution is correct \citep{brown2024large}.
In more general settings, the simplest and most well-established TTS method is \emph{Self-Consistency}, i.e.\ generating multiple independent samples and selecting an answer by plurality vote \citep{wang2022self}. This section evaluates ES fine-tuning in this role. The same math benchmarks are still used for this evaluation, however, instead of checking whether \emph{any} of the $k$ generated answers are correct, self-consistency is first used to identify one answer among the $k$ generated. The evaluation thus measures how well TTS can be used with ES-generated output distributions in non-verifiable test settings.

Self-Consistency was applied using the same models, parameters, and data as Section~\ref{sec:pass@k-math}.
The results mirror those in Section~\ref{subsec:math_experiments}: As $k$ increases, ES overtakes RL in terms of accuracy (Figures \ref{fig:self_cons_math500}, \ref{fig:self_cons_olympiad}, and \ref{fig:self_cons_minerva}). These results indicate that not only is ES more likely to include the correct answer in its distribution, but \emph{its maximum probability answer is more likely to be correct}. It is notable that on two of the three benchmarks the maximum probability answers of RL are worse than the base model, consistent with other observations of distribution degradation. Coupled with the pass@k results, these results suggest that the higher-quality distributions of ES may be generally more useful in TTS than those of RL. Exploring how ES can improve other forms of TTS, such as agentic harnesses \citep{ning2026code} or tree search \citep{aygun2026ai} is a promising avenue of future work.

\section{Future Work}
As the importance of TTS continues to grow, methods that explicitly optimize for use in TTS settings must be developed. ES is a natural fit given its gradient-free nature, which allows a TTS objective to be defined and optimized directly. For example, ES could be used to optimize the pass@k metric during training, directly aligning the training objective with the downstream metric that matters in TTS deployments. Beyond pass@k, this framework could extend to other TTS objectives such as self-consistency voting accuracy or multi-step agentic success rate.

\section{Conclusion}

As LLMs are increasingly used for discovery problems, it is important that post-training methods do not degrade solution coverage. This paper shows that while RL improves pass@1 accuracy, it does so at the cost of solution coverage through distribution collapse. In contrast, ES preserves broad solution coverage while simultaneously increasing pass@1 accuracy. The benefits of ES for test-time scaling are demonstrated through pass@k performance, output distribution analysis, and downstream test-time scaling experiments, where ES consistently outperforms RL. These results position ES as a promising alternative to RL for post-training in discovery problems such as science, math, and coding, and in other settings where solution diversity is critical.

\clearpage
\bibliography{biblio}
\bibliographystyle{colm2024_conference}

\newpage
\appendix
\section{Appendix}
\label{appendix}

Below, additional experimental details required to reproduce the results are presented, followed by additional experimental results, including pass@$k$, accuracy distribution analysis, and model regressions and progressions.

\subsection{Additional experiment details}
The additional details required to fully document the ES and RL training for GSM8K and MATH for Qwen2.5 and Qwen3 models are documented below.

\paragraph{Training hyperparameters.} Table~\ref{tab:verl-hparams} and Table~\ref{tab:es-hparams} list the training parameters used with the VERL library (RL) and ES-at-Scale library (ES), respectively, for GSM8K and MATH training. Among ES models trained on MATH, Qwen2.5-Math-7B uses a batch size of $512$ and a maximum response length of $3{,}000$ tokens. While Qwen2.5-14B and Qwen2.5-32B use batch size of $1{,}024$ and a max response length of $8{,}192$ tokens.

\begin{table}[h!]
    \centering
    \small
    \begin{tabular}{lc}
        \toprule
        \textbf{Hyperparameter} & \textbf{Value} \\
        \midrule
        Algorithm & GRPO \\
        Optimizer            & AdamW \\
        Learning rate        & 1e-6 \\
        LR schedule          & constant (no warmup) \\
        Train batch size     & $512$ \\
        Rollout batch size   & $1,024$ \\
        PPO mini-batch size  & $128$ \\
        PPO clip ratio $\varepsilon$           & $0.2$ \\
        Rollout $n$ (samples/prompt) & 8 \\
        Max prompt length    & $512$ \\
        Max response length  & $512$ \\
        Rollout temperature  & $1.0$ \\
        KL coefficient       & $0.0$ \\
        Clip ratio ($\epsilon$) & $0.2$ \\

        Total training steps & $500$ \\
        \bottomrule
    \end{tabular}
    \caption{VERL training hyperparameters for GSM8K.}
    \label{tab:verl-hparams}
\end{table}

\begin{table}[h!]
    \centering
    \small
    \begin{tabular}{lcc}
        \toprule
        \textbf{Hyperparameter} & \textbf{GSM8K} & \textbf{MATH} \\
        \midrule
        $\sigma$        & $0.001$  & $0.001$ \\
        $\alpha$         & $\alpha/2$ & $\alpha/2$ \\
        Train batch size     & $512$   & $512/1,024$ \\
        Population size & 32 & $32$ \\
        Max response length  & $512$  & $3,000/8,192$ \\
        Rollout temperature  & $0.0$   & $0.0$ \\
        Total training steps & $500$  & $500$ \\
        \bottomrule
    \end{tabular}
    \caption{Hyperparameters for training ES using the ES-at-Scale library.}
    \label{tab:es-hparams}
\end{table}

\paragraph{Training \& evaluation GPU infrastructure.}
The GSM8K experiments (Qwen2.5-Instruct-1.5B, -3B, -7B and Qwen3-1.7B, -4B, -8B) were trained for both ES and RL using 8 NVIDIA H200 GPUs, and each trained model was evaluated using a single NVIDIA H200 GPU. The MATH experiments (Qwen2.5-Math-7B, Qwen2.5-14B, and -32B) for ES were trained using 8 NVIDIA B200 GPUs, and each trained model was evaluated using a single NVIDIA B200 GPU.

\paragraph{Model chat templates.}
The Qwen chat template is used for the Qwen2.5 and Qwen3 models across all parameter scales. The wording of the system prompts differ slightly and so does the required answer format: for GSM8K the model must produce its final answer after ``\#\#\#\#'', whereas for MATH the model must place its final answer inside \verb|\boxed{}|. Table~\ref{table:appendix-qwen-template-gsm8k} documents the template used for GSM8K, and Table~\ref{table:appendix-qwen-template-math} documents the template used for MATH.

\begin{table*}
\centering
\small
\setlength{\tabcolsep}{6pt}
\renewcommand{\arraystretch}{1.3}
\begin{tabularx}{\linewidth}{|>{\centering\arraybackslash}m{2cm}|>{\raggedright\arraybackslash}X|}
\hline
\textbf{GSM8K} & \ttfamily <|im\_start|>system\textbackslash n You are a helpful AI assistant. Let's think step by step and output the final answer after ``\#\#\#\#''.<|im\_end|>\textbackslash n <|im\_start|>user\textbackslash n \textcolor{red}{\{question\}} <|im\_end|>\textbackslash n <|im\_start|>assistant\textbackslash n \\
\hline
\end{tabularx}
\caption{System prompt and template used for GSM8K.}
\label{table:appendix-qwen-template-gsm8k}
\end{table*}

\begin{table*}
\centering
\small
\setlength{\tabcolsep}{6pt}
\renewcommand{\arraystretch}{1.3}
\begin{tabularx}{\linewidth}{|>{\centering\arraybackslash}m{2cm}|>{\raggedright\arraybackslash}X|}
\hline
\textbf{MATH} & \ttfamily <|im\_start|>system\textbackslash n Please reason step by step, and put your final answer within \textbackslash boxed\{\}.<|im\_end|>\textbackslash n <|im\_start|>user\textbackslash n \textcolor{red}{\{question\}} <|im\_end|>\textbackslash n <|im\_start|>assistant\textbackslash n \\
\hline
\end{tabularx}
\caption{System prompt and template used for MATH.}
\label{table:appendix-qwen-template-math}
\end{table*}

\subsection{Additional experimental results}
\begin{figure*}[t!]
    \centering
    % --- Column headers ---
    \begin{subfigure}[b]{0.05\textwidth}
        \centering\phantom{X}
    \end{subfigure}\hfill
    \begin{subfigure}[b]{0.30\textwidth}
        \centering\textbf{MATH500}
    \end{subfigure}\hfill
    \begin{subfigure}[b]{0.30\textwidth}
        \centering\textbf{Olympiad Bench}
    \end{subfigure}\hfill
    \begin{subfigure}[b]{0.30\textwidth}
        \centering\textbf{Minerva}
    \end{subfigure}

    \vspace{0.5em}

    % --- Row 1: Qwen2.5-Math-7B ---
    \begin{subfigure}[c]{0.05\textwidth}
        \centering\rotatebox{90}{\textbf{Math-7B}}
    \end{subfigure}\hfill
    \begin{subfigure}[c]{0.30\textwidth}
        \centering
        \includegraphics[width=\textwidth]{plots/qwen2.5-math-7b/passk_qwen2_5_math_7b_math500_evalTemp0.6.pdf}
    \end{subfigure}\hfill
    \begin{subfigure}[c]{0.30\textwidth}
        \centering
        \includegraphics[width=\textwidth]{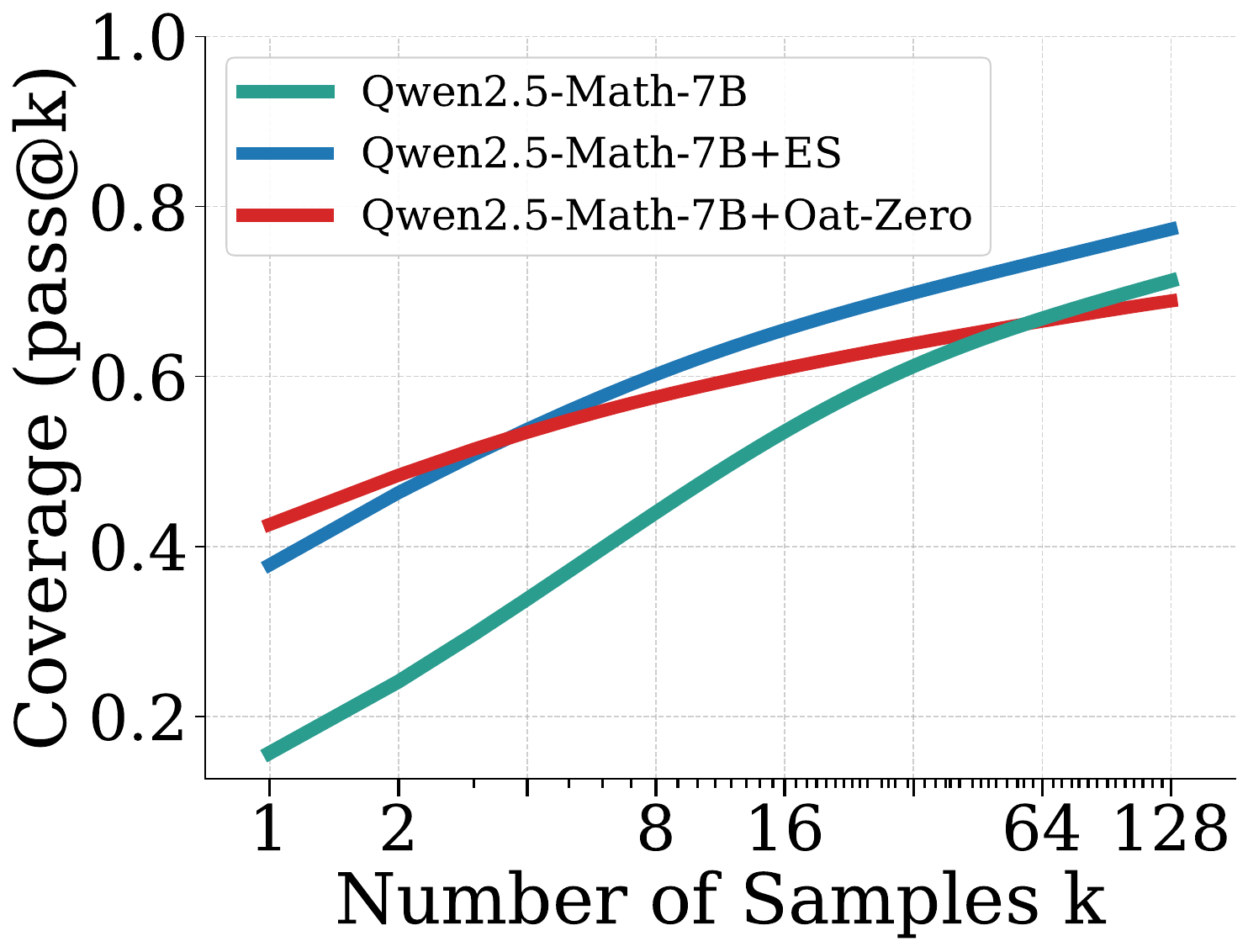}
    \end{subfigure}\hfill
    \begin{subfigure}[c]{0.30\textwidth}
        \centering
        \includegraphics[width=\textwidth]{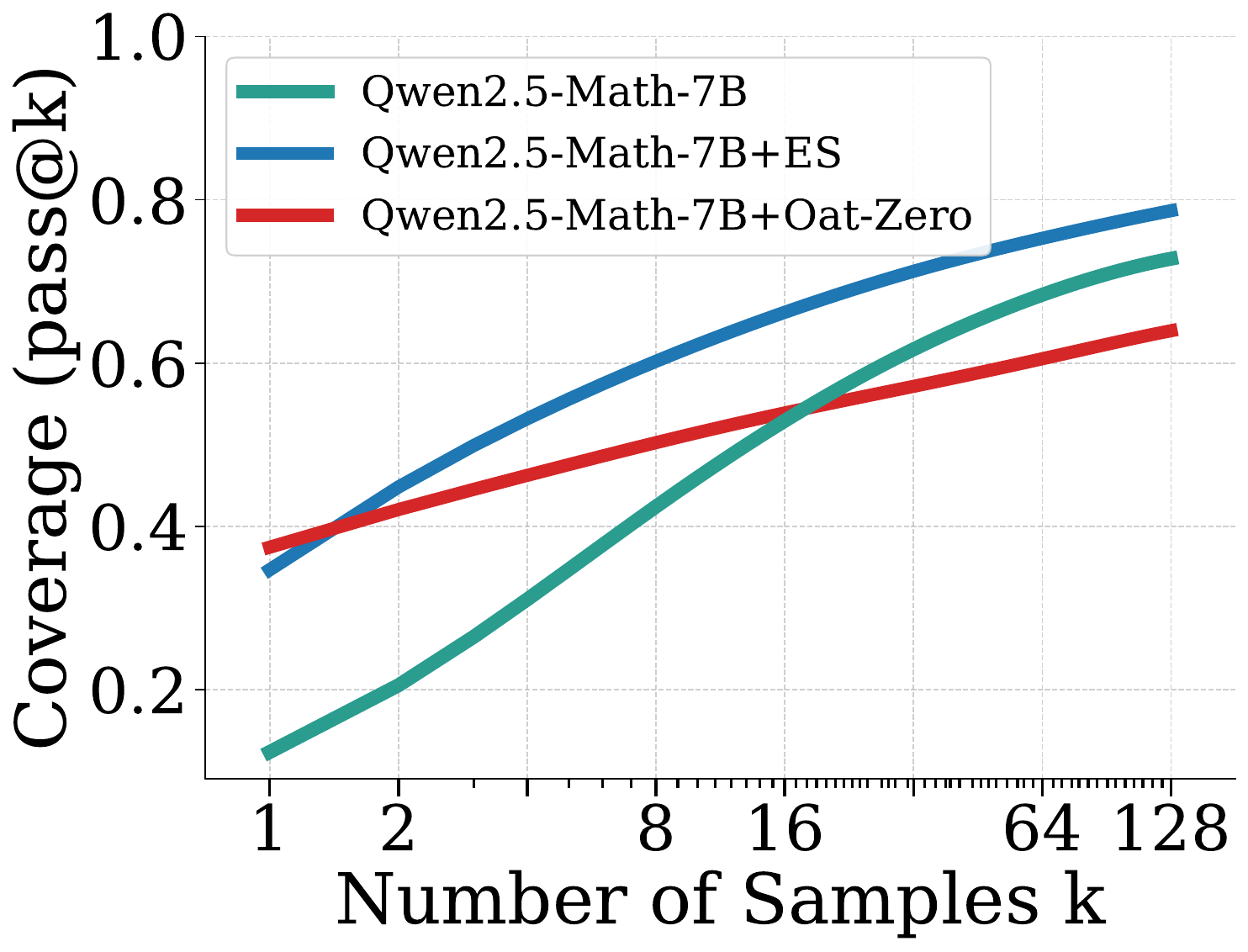}
    \end{subfigure}

    \vspace{0.5em}

    % --- Row 2: Qwen2.5-14B ---
    \begin{subfigure}[c]{0.05\textwidth}
        \centering\rotatebox{90}{\textbf{14B}}
    \end{subfigure}\hfill
    \begin{subfigure}[c]{0.30\textwidth}
        \centering
        \includegraphics[width=\textwidth]{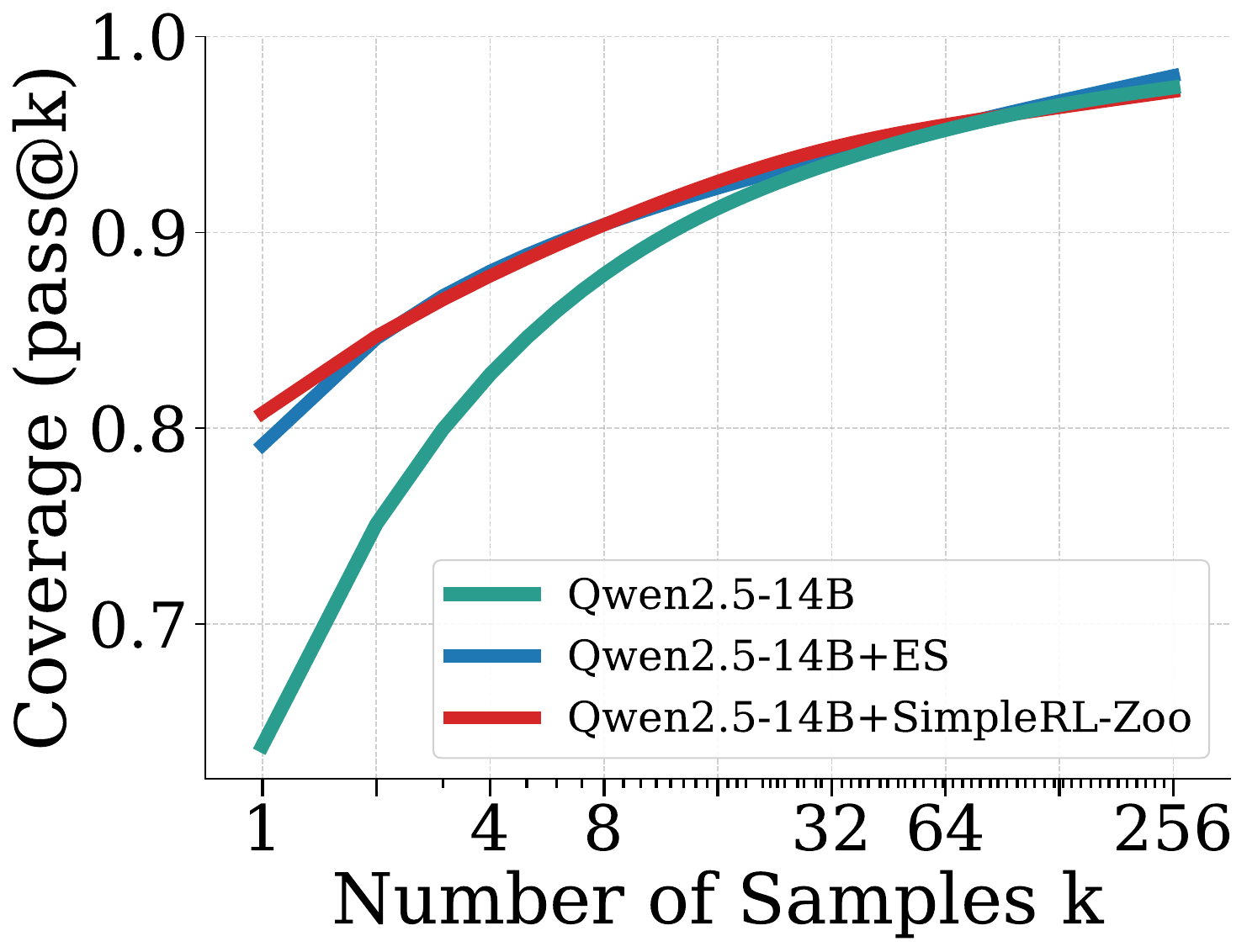}
    \end{subfigure}\hfill
    \begin{subfigure}[c]{0.30\textwidth}
        \centering
        \includegraphics[width=\textwidth]{plots/qwen2.5-14b/passk_qwen2_5_14b_olympiad_bench_evalTemp0.6.pdf}
    \end{subfigure}\hfill
    \begin{subfigure}[c]{0.30\textwidth}
        \centering
        \includegraphics[width=\textwidth]{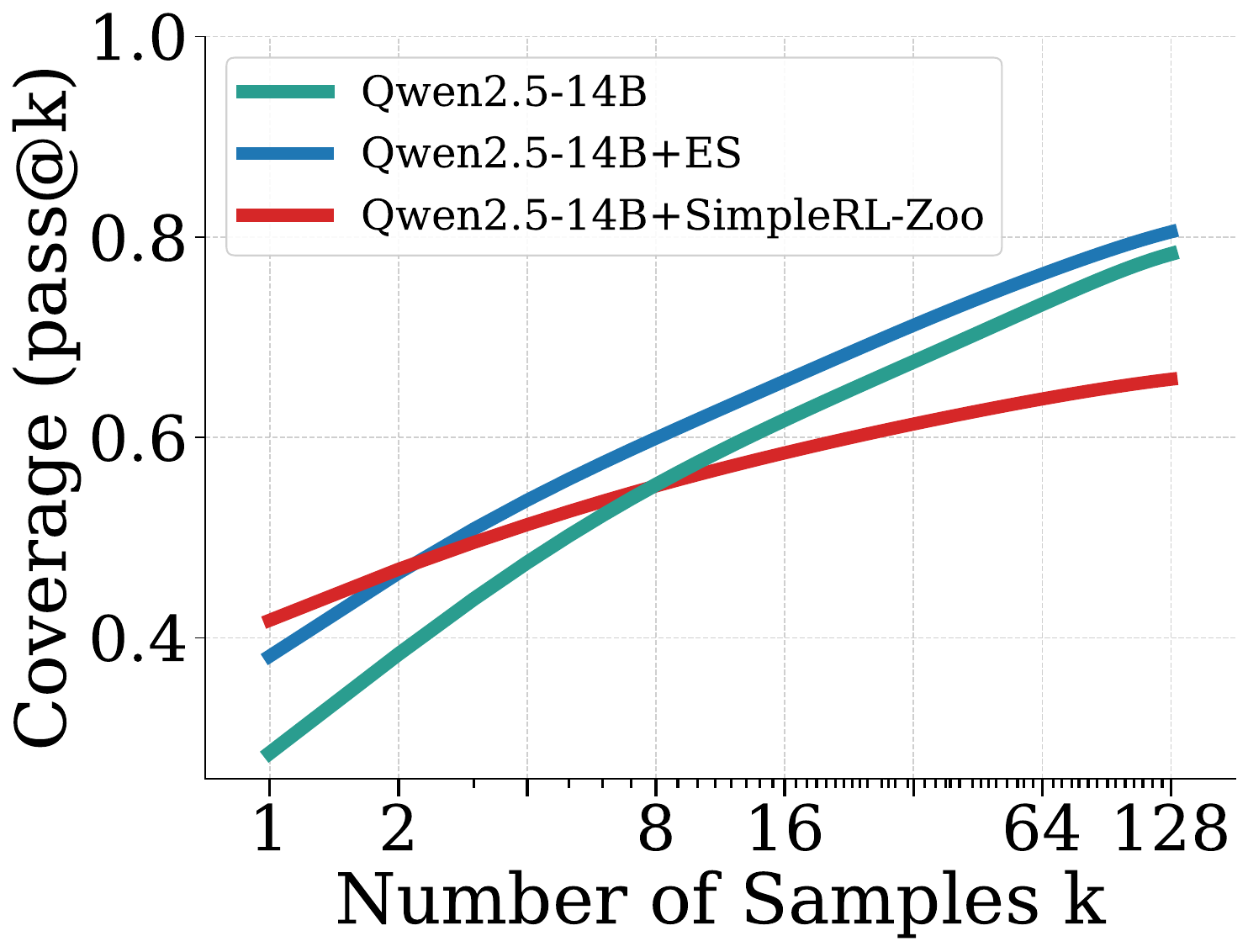}
    \end{subfigure}

    \vspace{0.5em}

    % --- Row 3: Qwen2.5-32B ---
    \begin{subfigure}[c]{0.05\textwidth}
        \centering\rotatebox{90}{\textbf{32B}}
    \end{subfigure}\hfill
    \begin{subfigure}[c]{0.30\textwidth}
        \centering
        \includegraphics[width=\textwidth]{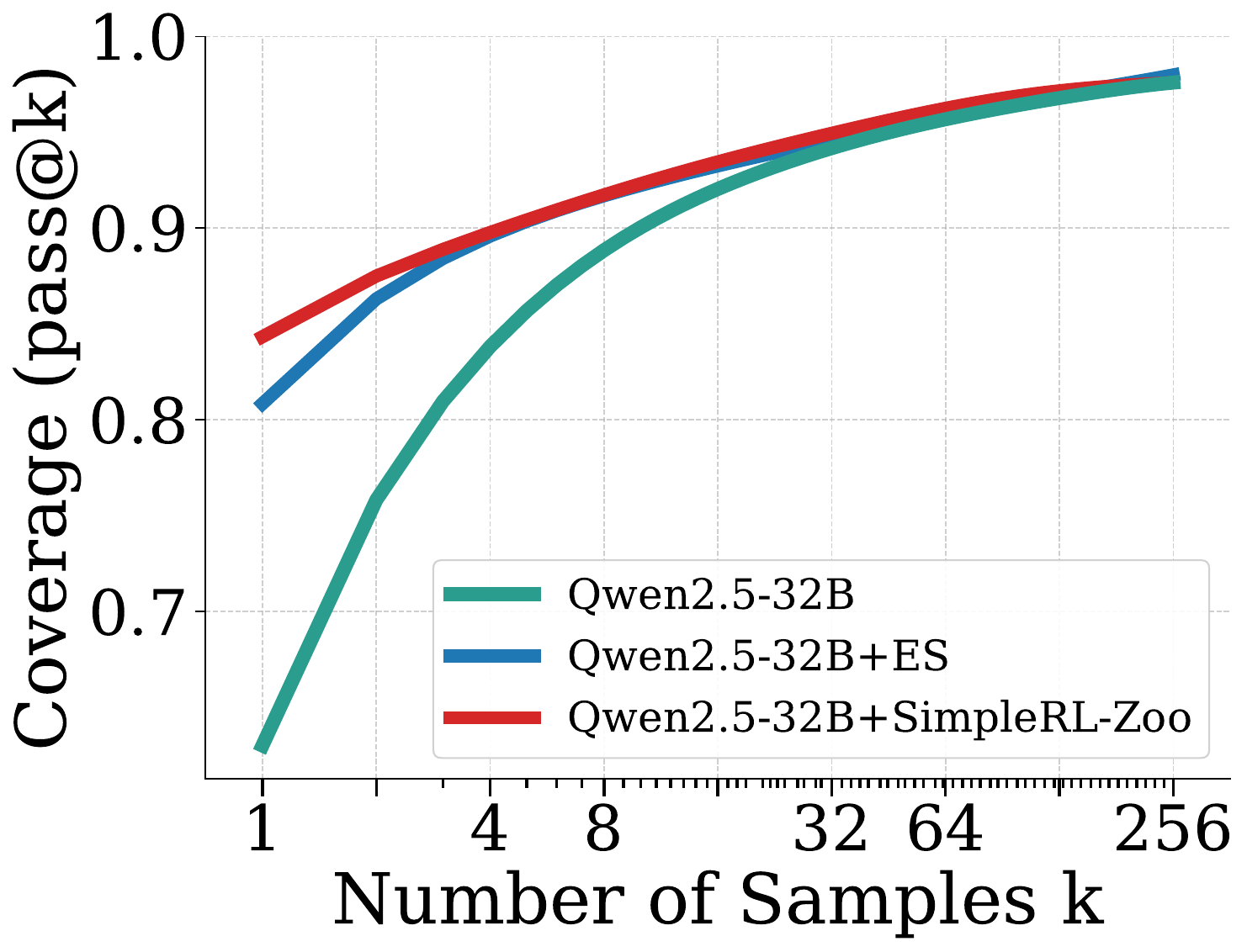}
    \end{subfigure}\hfill    
    \begin{subfigure}[c]{0.30\textwidth}
        \centering
        \includegraphics[width=\textwidth]{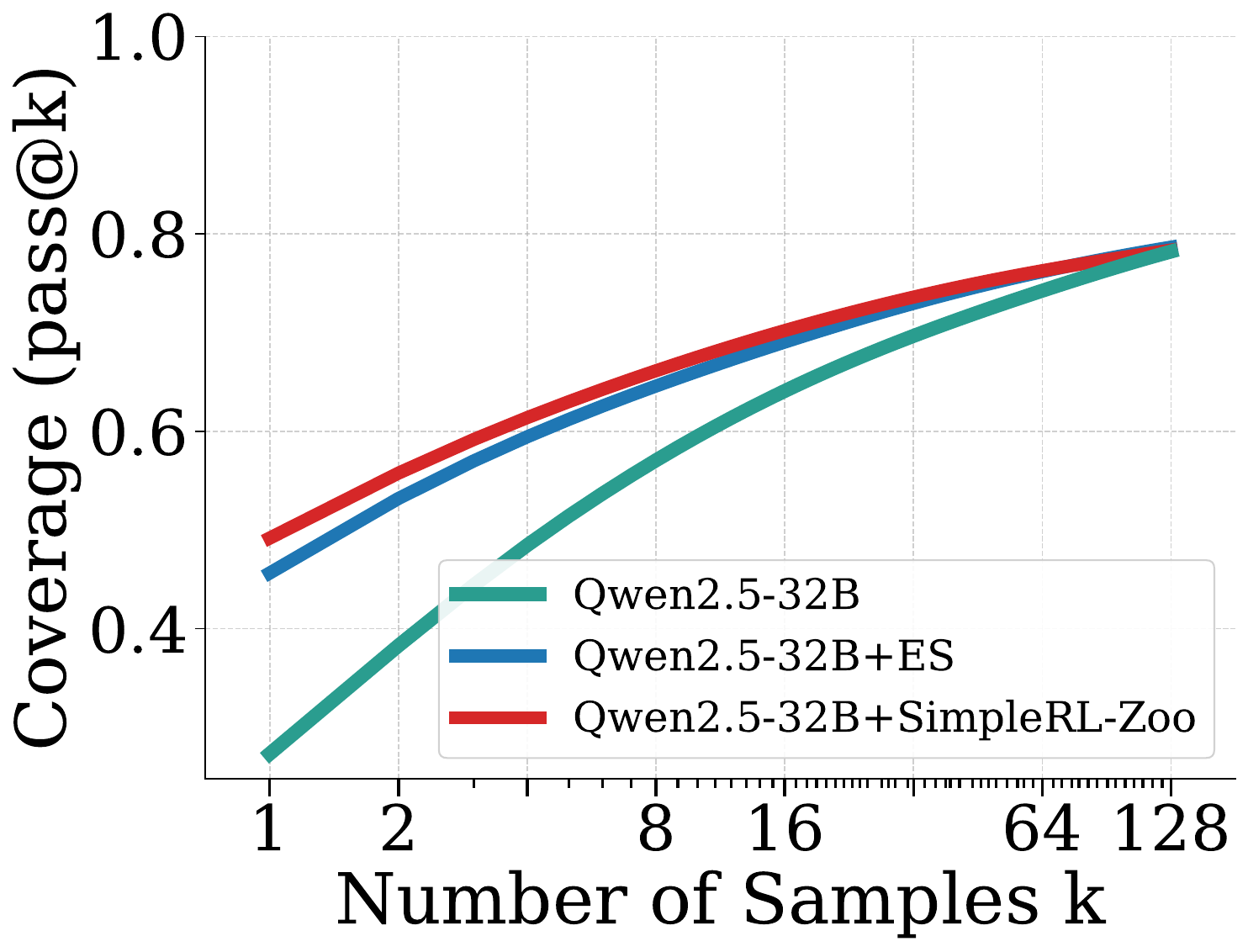}
    \end{subfigure}\hfill
    \begin{subfigure}[c]{0.30\textwidth}
        \centering
        \includegraphics[width=\textwidth]{plots/qwen2.5-32b/passk_qwen2_5_32b_minerva_evalTemp0.6.pdf}
    \end{subfigure}

    \caption{\emph{Qwen2.5 results for MATH500, Olympiad Bench and Minerva across Math-7B, 14B and 32B model scales.} ES outperforms SOTA RL training recipes (OatZero and SimpleRL-Zoo) for pass@$k$, highlighting the pass@$k$ trend in favor of ES as model parameters scale. This further motivates the use of ES for problems requiring TTS with larger models.}
    \label{fig:appendix-qwen2.5-passk}
\end{figure*}

This section includes additional experiment results for pass@k results for MATH, additional accuracy analysis and regressions and progressions for Qwen2.5 and Qwen3 models ranging from 1.5B to 32B parameters across GSM8K, Minerva, Olympaid Bench, and MATH500,

\paragraph{Pass@k for MATH experiments}
Figure~\ref{fig:appendix-qwen2.5-passk} presents additional pass@k results for Qwen2.5-Math-7B, Qwen2.5-14B, and Qwen2.5-32B on MATH500, Olympiad Bench, and Minerva. The trends from Section~\ref{sec:pass@k-performance} hold at larger scales: ES outperforms RL on pass@k for Qwen2.5-Math-7B and Qwen2.5-14B across every benchmark, underscoring the quality of the ES distribution as parameter count grows. For Qwen2.5-32B, ES remains competitive with RL on MATH500 and Olympiad Bench and outperforms it on the challenging Minerva benchmark. Notably, the base model never surpasses the ES fine-tuned model at any scale, confirming that ES preserves broad solution coverage even as parameter count increases.

\begin{figure}
    \vspace*{1ex}
    \centering
    \includegraphics[width=1.0\columnwidth]{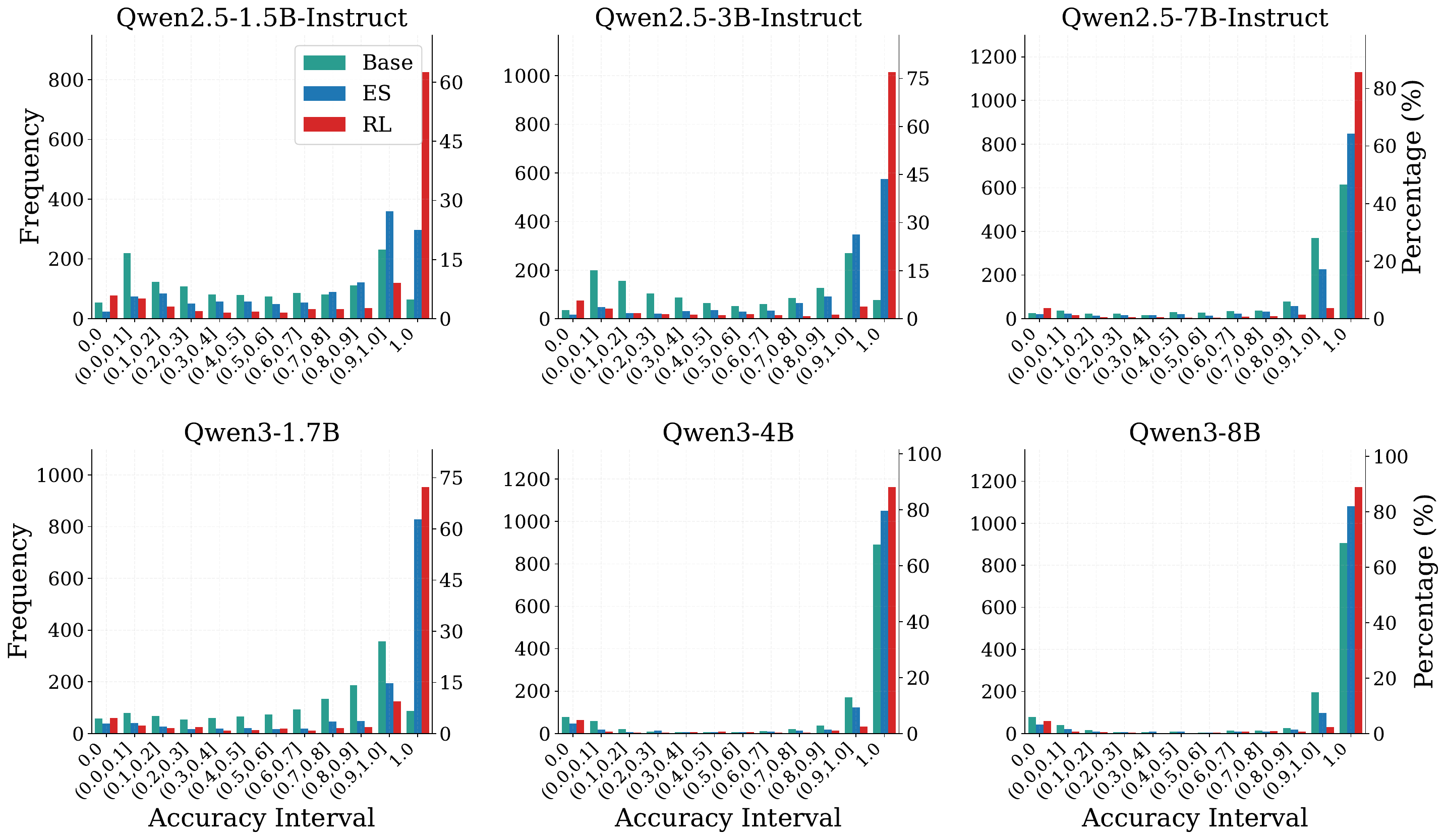}
    \caption{\emph{Accuracy distribution analysis for Qwen2.5 and Qwen3 models from 1.5B to 8B on GSM8K}.}
    \label{fig:appendix-gsm8k-accuracy-distribution}
    \vspace*{8ex}
\end{figure}

\begin{figure}
    \centering
    \includegraphics[width=1.0\columnwidth]{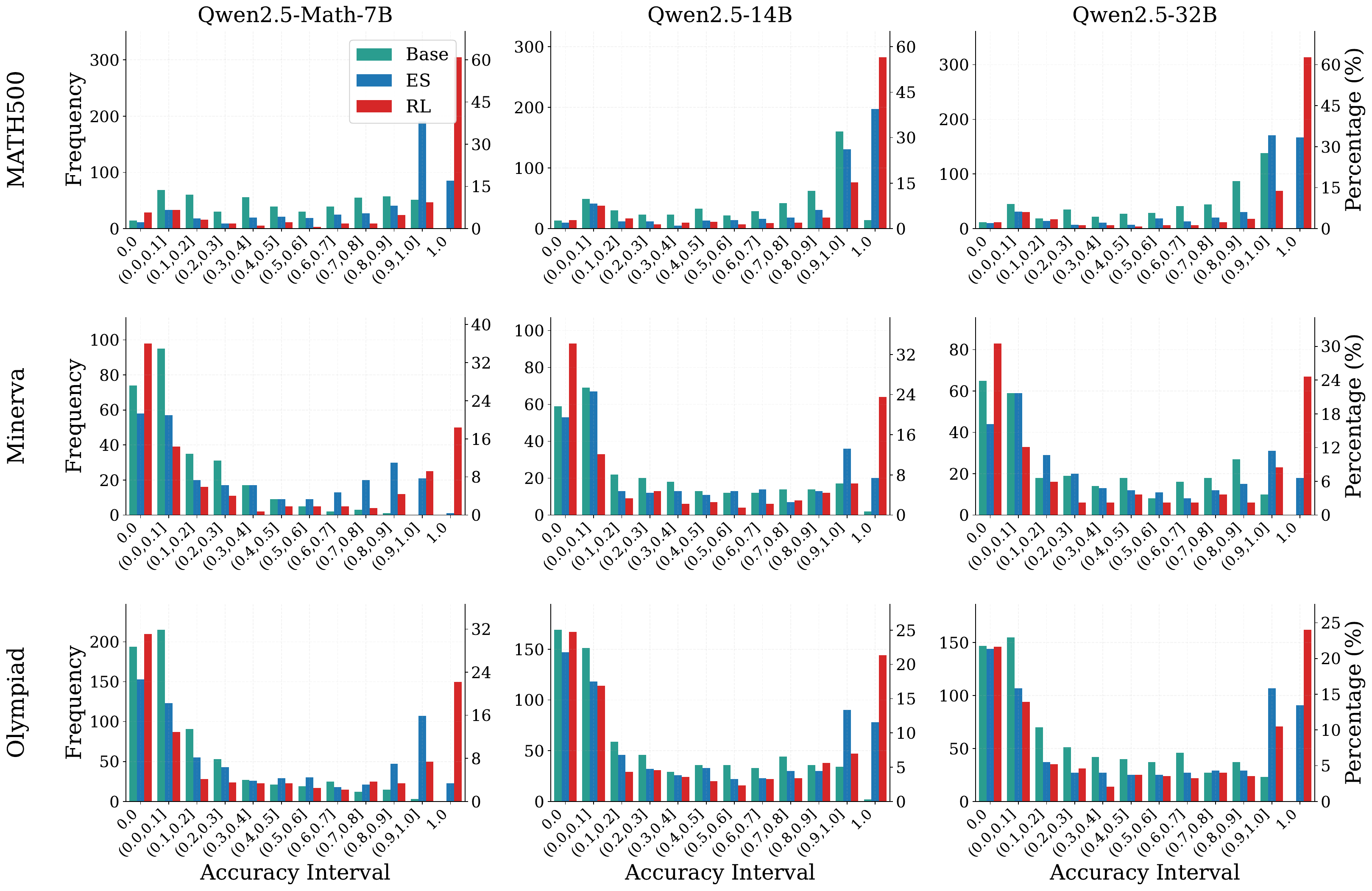}
    \caption{\emph{Accuracy distribution analysis for Qwen2.5 model from 7B to 32B on MATH500, Minerva, and Olympiad Bench}. ES reduces the number of unsolveable problems i.e., prompts located in bin 0.0 across each benchmark.}
    \label{fig:appendix-math-accuracy-distribution}
\end{figure}

\paragraph{Accuracy distribution analysis.}
Figures \ref{fig:appendix-gsm8k-accuracy-distribution} and \ref{fig:appendix-math-accuracy-distribution} present accuracy distribution plots for ES and RL across all trained models, extending the results in Section~\ref{sec:solution-coverage} and Figure~\ref{fig:accuracy-dist}. As in Section~\ref{sec:pass@k-math}, ES consistently reduces the number of prompts in bin 0.0 relative to the base model, while RL increases this count. Figure~\ref{fig:appendix-gsm8k-accuracy-distribution} shows this pattern for Qwen2.5-Instruct and Qwen3 models (1.5B--8B parameters) trained on GSM8K. Figure~\ref{fig:appendix-math-accuracy-distribution} extends this to ES and RL recipes (\emph{OatZero}, \emph{SimpleRL-Zoo}) for Qwen2.5-Math-7B, Qwen2.5-14B, and Qwen2.5-32B on MATH500, Olympiad Bench, and Minerva, showing the same bin 0.0 pattern across every model-benchmark pair. Notably, RL increases the number of prompts in bin 1.0 with respect to ES and the base model. Whereas ES's gains concentrate in the $(0.9, 1.0]$ range. These results further explain the pass@k findings in Figure~\ref{fig:appendix-qwen2.5-passk}. For test-time scaling, this distinction matters directly: a lower count in bin 0.0 means ES yields fewer prompts that are entirely unsolvable regardless of how many samples $k$ are drawn, while RL's increase in unsolvable prompts imposes a hard ceiling on achievable pass@k performance that no amount of additional sampling can overcome.

\begin{figure}
    \centering
    \includegraphics[width=1.0\linewidth]{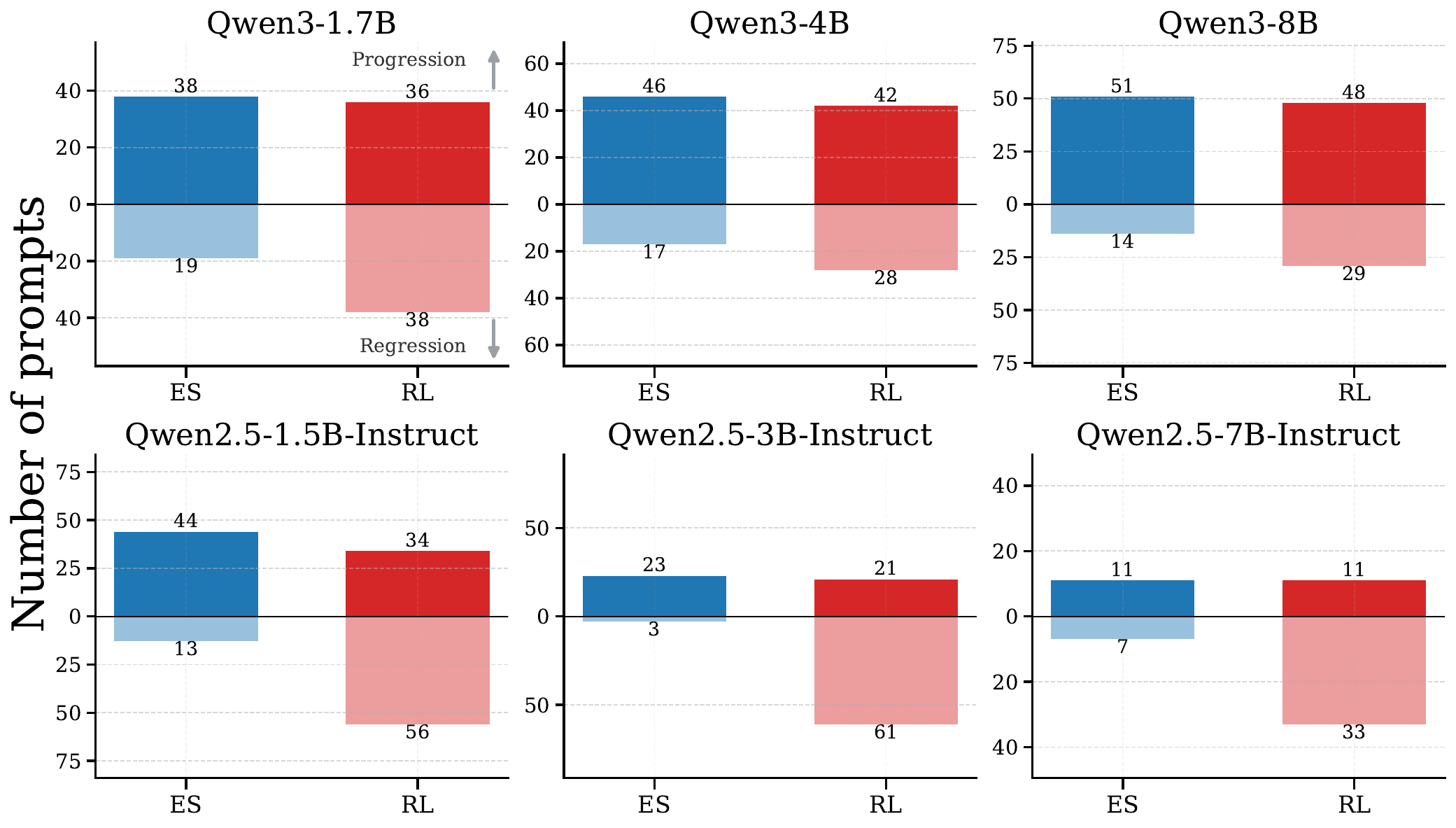}
    \caption{\emph{Progressions and regression for Qwen2.5 and Qwen3 models trained on GSM8K with parameters rangin from 1.5B to 8B}. ES reduces the number of regressions and increases progressions compared to RL across all parameter scales.}
    \label{fig:appendix-progressions-regressions-gsm8k}
\end{figure}

\begin{figure}
    \centering
    \includegraphics[width=1.0\linewidth]{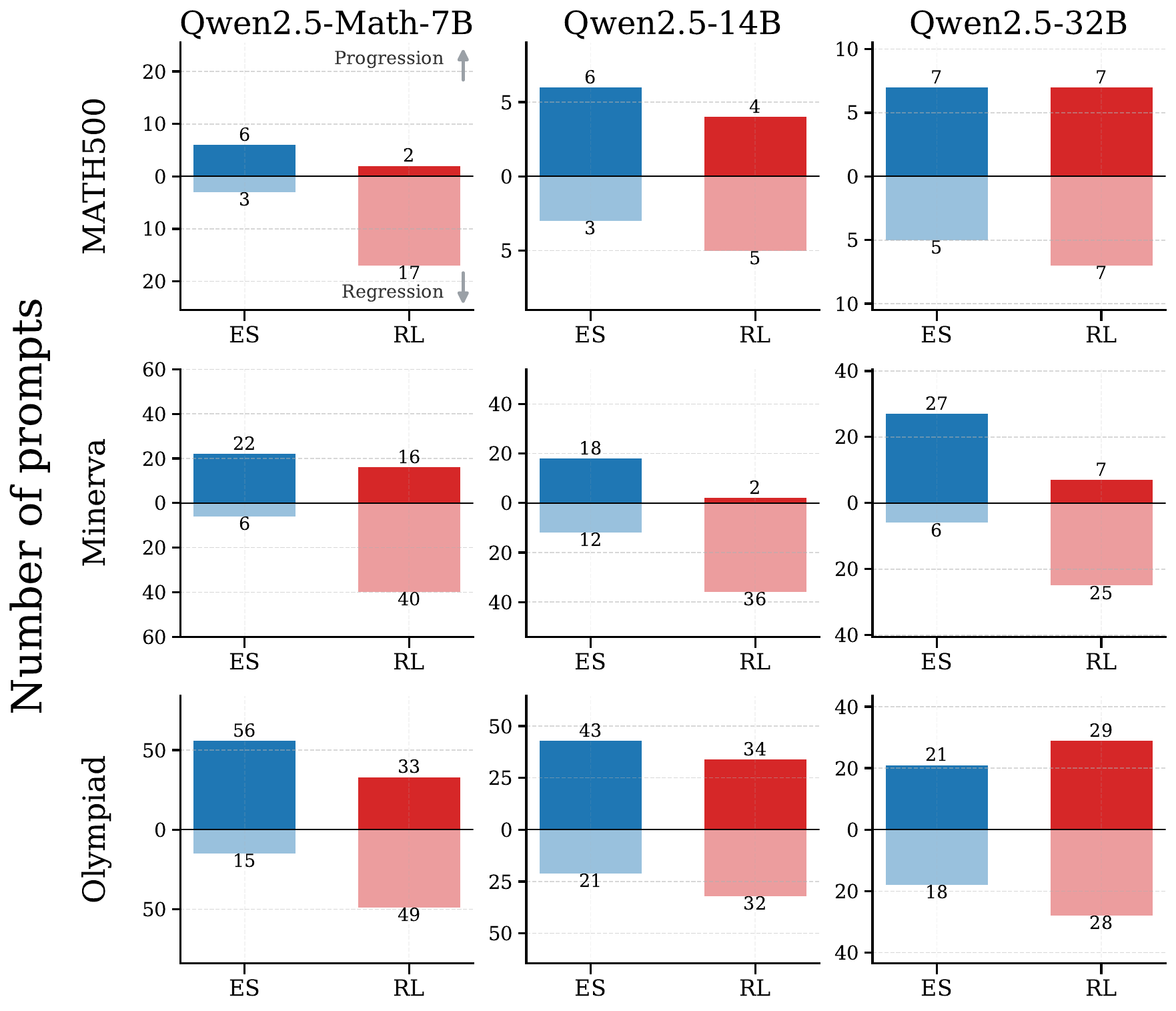}
    \caption{\emph{Progressions and regression for Qwen2.5-Math-7B, Qwen2.5-14B, and Qwen2.5-32B trained on MATH and evaluated using MATH500, Olympiad Bench, and Minerva}. RL has more regressions compared to ES across each benchmark, while ES has more progressions.}
    \label{fig:appendix-progressions-regressions-math}
    \vspace*{35ex}
\end{figure}

\paragraph{Measuring model regressions and progressions.}
Figures \ref{fig:appendix-progressions-regressions-gsm8k} and \ref{fig:appendix-progressions-regressions-math} present model progressions and regressions for ES and RL fine-tuned models on GSM8K and MATH across all model parameter scales from 1.5B to 32B. Figure \ref{fig:appendix-progressions-regressions-gsm8k} shows that ES increases model progressions and reduces model regressions compared to RL on GSM8K for Qwen2.5-Instruct and Qwen3 models from 1.5B to 8B parameters. Figure \ref{fig:appendix-progressions-regressions-math} shows the same pattern on MATH across MATH500, Olympiad Bench, and Minerva for Qwen2.5-Math-7B, Qwen2.5-14B, and Qwen2.5-32B, comparing ES against RL recipes \emph{OatZero} and \emph{SimpleRL-Zoo}. Together, Figures \ref{fig:appendix-progressions-regressions-gsm8k} and \ref{fig:appendix-progressions-regressions-math} show that ES's advantage in progressions and regressions holds consistently across model families, benchmarks, and parameter scales, reinforcing that ES is less destructive to existing model knowledge than RL regardless of scale. This scale-invariance of ES's regression/progression advantage complements the pass@k results in Section~\ref{sec:pass@k-performance}, further supporting ES as a more knowledge-preserving post-training method than RL. Fewer regressions directly motivate the use of ES in test-time scaling settings. ES maintains a larger pool of correct solutions to draw on when scaling test-time compute, whereas RL's regressions permanently remove some of these solutions from consideration regardless of how many samples are drawn.

\end{document}